\documentclass[runningheads]{llncs}
\usepackage{graphicx}
\usepackage{amsmath,amssymb} 
\usepackage{times}
\usepackage{epsfig}
\usepackage{math}
\usepackage{algorithm}
\usepackage{algorithmic}
\usepackage{xcolor}
\usepackage{multirow}
\usepackage[lofdepth,lotdepth]{subfig}
\usepackage{booktabs}

\begin{document}
\title{DeepGUM: Learning Deep Robust Regression with a Gaussian-Uniform Mixture Model}

\titlerunning{DeepGUM: Learning Deep Robust Regression}

\authorrunning{S. Lathuili\`{e}re et al.}

%
\author{St\'{e}phane~Lathuili\`{e}re\inst{1,3},
  Pablo~Mesejo\inst{1,2},
  Xavier~Alameda-Pineda\inst{1}, 
  and~Radu~Horaud\inst{1}}
%

\authorrunning{S. Lathuili\`{e}re et al.}

\institute{Inria Grenoble Rh\^{o}ne-Alpes, Montbonnot-Saint-Martin, France, \and
  University of Granada, Granada, Spain,\and
  University of Trento, Trento, Italy\\
  \email{ {firstname.name}@inria.fr}}

\maketitle              
\begin{abstract}
In this paper we address the problem of how to robustly train a ConvNet for regression, or deep robust regression.
Traditionally, deep regression employ the $L_2$ loss function, known to be sensitive to outliers, i.e. samples that either lie at 
an abnormal distance away from the majority of the training samples, or that correspond to wrongly annotated targets. This means that, during back-propagation, outliers may bias the training process due to the high magnitude of their gradient. In this paper, we 
propose DeepGUM: a deep regression model that is robust to outliers thanks to the use of a 
Gaussian-uniform mixture model. We derive an optimization algorithm that alternates between the unsupervised  detection of outliers 
using expectation-maximization, and the supervised training with \textit{cleaned} samples using stochastic 
gradient descent. DeepGUM is able to adapt to a continuously evolving outlier distribution, avoiding to manually impose any threshold on the 
proportion of outliers in the training set. Extensive experimental evaluations on four different tasks (facial and fashion landmark detection, age and head pose estimation) lead us to conclude that our novel robust technique provides reliability in the presence of various types of noise and protection against a high percentage of outliers. 

\end{abstract}


\section{Introduction}
\label{sec:intro}
For the last decade, deep learning architectures have undoubtably established the state of the art in computer vision tasks such as image classification~\cite{Krizhevsky2012,Szegedy2015} or object 
detection~\cite{Girshick2014,Sermanet2014}. These architectures, e.g. ConvNets, consist of 
several convolutional 
layers, followed by a few fully connected layers and by a classification softmax layer with, for instance, a cross-entropy 
loss. 
ConvNets have also been used for regression, i.e. predict continuous as opposed to 
categorical output values. Classical regression-based computer vision methods have addressed human pose estimation~\cite{Toshev2014}, age 
estimation~\cite{rothe2016deep}, head-pose estimation~\cite{Demirkus:2015}, or facial landmark detection~\cite{Sun2013}, 
to cite a few. Whenever ConvNets are used for learning a regression network, the softmax layer is replaced with a fully 
connected layer, with linear or sigmoid activations, and $L_2$ is often used to measure the discrepancy 
between prediction and target variables. It is well known that $L_2$-loss is
strongly sensitive to outliers, potentially leading to poor generalization performance~\cite{huber2004}. While robust regression is extremely well investigated in statistics, there has only been a handful of methods that combine robust regression with deep architectures.

\begin{figure}[t!]\centering
\includegraphics[width=0.9\linewidth]{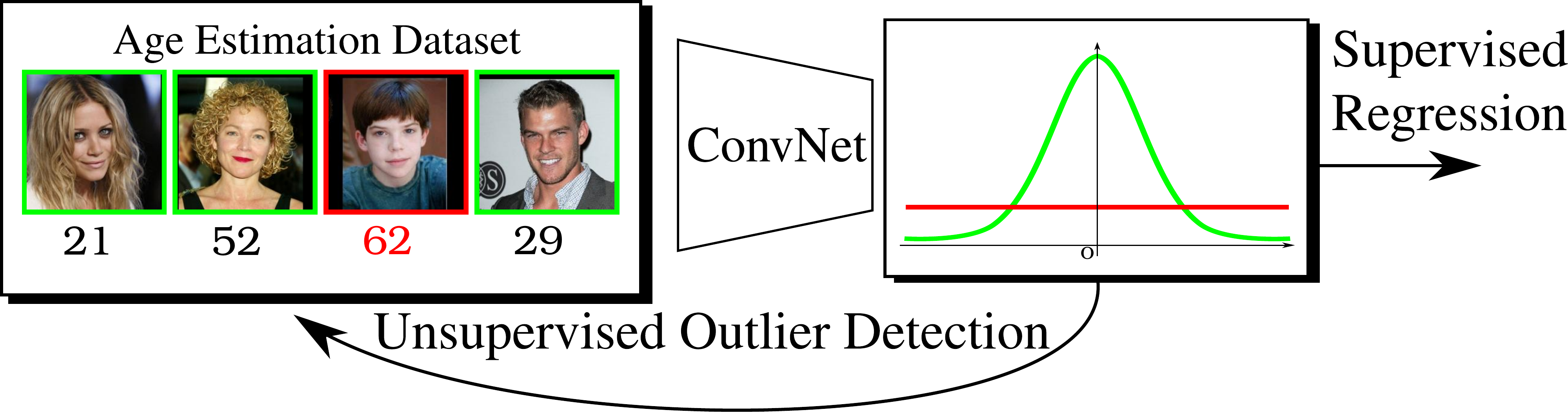}
\caption{A Gaussian-uniform mixture model is combined with a ConvNet architecture to downgrade the influence of wrongly annotated targets (outliers) on the learning process.
}
\label{fig:teaser}
\end{figure}

This paper proposes to mitigate the influence of outliers when deep neural architectures are used to learn a regression function, ConvNets in particular. More precisely, we investigate a methodology specifically designed to cope with two types of outliers that are often encountered: (i) samples  that lie at an abnormal distance away from the other training samples, and (ii) wrongly annotated training samples. On the one hand, abnormal samples are present in almost any measurement system and they are known to bias the regression parameters. 
On the other hand, deep learning requires very large amounts of data and the annotation process, be it either automatic or manual, is inherently prone to errors. These unavoidable issues fully justify the development of robust deep regression.

The proposed method combines the representation power of ConvNets with the principled probabilistic mixture framework for outlier detection and rejection, e.g. Figure~\ref{fig:teaser}. We propose to use a Gaussian-uniform mixture (GUM) as the last layer of a ConvNet, and we refer to this combination as DeepGUM. The mixture model hypothesizes a Gaussian distribution for inliers and a uniform distribution for outliers. We interleave an EM procedure within stochastic gradient descent (SGD) to downgrade the influence of outliers in order to robustly estimate the network parameters. 
We empirically validate the effectiveness of the proposed method with four computer vision problems and associated datasets: facial and fashion landmark detection, age estimation, and head pose estimation. The standard regression measures are accompanied by statistical tests that discern between random differences and systematic improvements.
%

The remainder of the paper is organized as follows. Section~\ref{sec:related_work} describes the related work. Section~\ref{sec:model} describes in detail the proposed method and the associated algorithm. Section~\ref{expe} describes extensive experiments with several applications and associated datasets. Section~\ref{sec:conclusions} draws conclusions and discusses the potential of robust deep regression in computer vision.

\section{Related Work}
\label{sec:related_work}


Robust regression has long been studied in statistics \cite{huber2004,maronna2006robust,rousseeuw2005robust} 
and in computer vision \cite{black1996unification,meer1991robust,stewart1999robust}. 
Robust regression methods have a high \textit{breakdown point}, which is the smallest amount of 
outlier contamination that an estimator can handle before yielding poor results. Prominent examples are the least trimmed squares, the Theil-Sen estimator or heavy-tailed 
distributions~\cite{gelman2003}. Several robust training strategies for artificial neural 
networks are also available~\cite{Beliakov2011,Neuneier1998}.

M-estimators, sampling methods, trimming methods and 
robust clustering are among the most used robust statistical methods. M-estimators~\cite{huber2004} minimize the sum of a positive-definite function of the residuals and attempt to reduce the influence of large residual values. The minimization is carried our with weighted least 
squares techniques, with no proof of convergence for most M-estimators. Sampling 
methods~\cite{meer1991robust}, such as least-median-of-squares or random sample consensus (RANSAC), estimate the 
model parameters by solving a system of equations defined for a randomly chosen data subset. The main drawback of 
sampling methods is that they require complex data-sampling procedures and it is tedious to use them for estimating a 
large number of parameters. Trimming methods~\cite{rousseeuw2005robust} rank the residuals and down-weight the data points 
associated with large residuals. They are typically cast into a (non-linear) weighted least squares optimization problem, 
where the weights are modified at each iteration, leading to iteratively re-weighted least squares problems. Robust 
statistics have also been addressed in the framework of mixture models and a number of robust mixture models 
were proposed, such as Gaussian mixtures with a uniform noise 
component~\cite{banfield1993model,coretto2016robust}, heavy-tailed distributions~\cite{forbes2014new}, trimmed 
likelihood estimators~\cite{galimzianova2015robust,neykov2007robust}, or weighted-data mixtures \cite{gebru2016algorithms}. Importantly, it has been recently reported that modeling outliers with
an uniform component yields very good performance~\cite{coretto2016robust,gebru2016algorithms}.

Deep 
robust classification was recently addressed, e.g. \cite{Bekker2016} assumes that observed 
labels are generated from true labels with unknown noise parameters: a
probabilistic model that maps true labels onto observed labels is proposed and an EM 
algorithm is derived. In \cite{xiao2015learning} is proposed a probabilistic model that exploits the 
relationships 
between classes, images and noisy labels for large-scale image classification. This framework requires a dataset with explicit clean- and noisy-label annotations as well as an additional dataset annotated with a noise type for 
each sample, thus making the method difficult to use in practice. Classification algorithms based on a distillation process to learn from noisy data was recently proposed \cite{li2017learning}.

Recently, deep regression methods were proposed, e.g.~\cite{Mukherjee2015,Ranjan2016,Sun2013,Toshev2014,Lathuiliere2017}.
Despite the vast robust statistics literature and the importance of regression in computer vision, at the best of our knowledge there has been only one attempt to combine robust regression with deep networks \cite{Belagiannis2015}, where robustness is achieved by minimizing the Tukey's bi-weight loss function, i.e. an M-estimator.
In this paper we take a radical different approach and propose to use robust mixture modeling within a ConvNet. We conjecture
that while \textit{inlier noise} follows a Gaussian distribution, \textit{outlier errors} are uniformly distributed over the volume occupied by the data. Mixture modeling provides a principled way to characterize data points individually, based on posterior probabilities. We propose an algorithm that interleaves a robust mixture model with network training, i.e. alternates between EM and SGD. EM evaluates data-posterior probabilities which are then used to weight the residuals used by the network loss function and hence to downgrade the influence of samples drawn from the uniform distribution. Then, the network parameters are updated which in turn are used by EM. A prominent feature of the algorithm is that it requires neither annotated outlier samples nor prior information about their percentage in the data. This is in contrast with \cite{xiao2015learning} that requires explicit inlier/outlier annotations and with \cite{Belagiannis2015} which uses a fixed hyperparameter ($c=4.6851$) that allows to exclude from SGD samples with high residuals.

\section{Deep Regression with a Robust Mixture Model}
\label{sec:model}

We assume that the inlier noise follows a Gaussian distribution while the 
outlier error follows a uniform distribution. Let $\high\in\mathbb{R}^M$ and $\low\in\mathbb{R}^D$ be the input image and 
the output vector with dimensions $M$ and $D$, respectively, with
$D\ll M$.
Let $\phivect$ denote a ConvNet with parameters $\wvect$ such that $\low=\phivect(\high,\wvect)$.
We aim to train a model that detects outliers and downgrades their role in the prediction of a network output, while there is no prior information about the percentage and spread of outliers.
The probability of $\low$ conditioned by $\high$ follows a Gaussian-uniform mixture model (GUM):
\begin{align}
  p(\low| \high;\thetavect,\wvect)  =  
   \pi \;\mathcal{N}(\low;\phivect(\high ; \wvect),\Sigmamat) +(1-\pi) \; \mathcal{U}(\low;\gamma),
  \label{eq:target-prob}
\end{align}
%
%
where 
$\pi$ is the prior probability of an inlier sample, 
$\gamma$ is the normalization parameter
of the uniform distribution and $\Sigmamat\in\mathbb{R}^{D\times D}$ is the covariance matrix of the multivariate Gaussian distribution. Let
$\thetavect=\{\pi,\gamma,\Sigmamat\}$ be the parameter set of GUM. At training we estimate the parameters of the mixture model, $\thetavect$, and of the network, $\wvect$. An EM algorithm is used to estimate the former together with the responsibilities $r_n$, which are plugged into the network's loss, minimized using SGD so as to estimate the later.



\subsection{EM Algorithm}
\label{sec:em}
Let a training dataset consist of $N$ image-vector pairs $\{\high_n,\low_n\}_{n=1}^N$.
At each iteration, EM alternates between evaluating the expected complete-data log-likelihood (E-step) and updating the parameter set $\thetavect$ conditioned by the network parameters (M-step). In practice, the E-step evaluates the posterior probability (responsibility) of an image-vector pair $n$ to be an inlier:
\begin{equation}
\label{eq:rn}
r_{n}(\thetavect^{(i)}) = \frac{\pi^{(i)} \mathcal{N}(\lown;\phivect(\highn,\wvect^{(c)}),\Sigmamat^{(i)})}{\pi^{(i)} 
\mathcal{N}(\lown;\phivect(\highn,\wvect^{(c)}),\Sigmamat^{(i)})+(1-\pi^{(i)})\gamma^{(i)}},
\end{equation}
where $(i)$ denotes the EM iteration index and $\wvect^{(c)}$ denotes the currently estimated network parameters. The posterior probability of the $n$-th data pair to be an outlier is $1-r_n(\thetavect^{(i)})$. The M-step updates the mixture parameters $\thetavect$ with:
\begin{align}
\label{eq:sigmamat} 
 \Sigmamat^{(i+1)} &= \sum_{n=1}^N 
r_{n}(\thetavect^{(i)}) \deltavect_n^{(i)}\deltavect_n^{(i)\top},\\
\label{eq:pi} 
 \pi^{(i+1)} & =\sum_{n=1}^Nr_{n}(\thetavect^{(i)})/N,\\
  \label{eq:lambda}
 \frac{1}{\gamma^{(i+1)}} &= \prod_{d=1}^{D} 2\sqrt{3\left(C^{(i+1)}_{2d}-\left(C^{(i+1)}_{1d}\right)^2\right)},
\end{align}
where $\deltavect_n^{(i)} = \low_n-\phivect(\high_n;\wvect^{(c)})$,  
and  $C_1$ and $C_2$ are the first- and second-order centered data moments computed using 
($\delta_{nd}^{(i)}$ denotes the $d$-th entry of $\deltavect_n^{(i)}$): 
 \begin{align}
      C_{1d}^{(i+1)}=\frac{1}{N}\sum_{n=1}^N\frac{(1-r_{n}(\thetavect^{(i)}))}{1- \pi^{(i+1)} }\delta_{nd}^{(i)},\;
      C_{2d}^{(i+1)}=\frac{1}{N}\sum_{n=1}^N\frac{(1-r_{n}(\thetavect^{(i)}))}{1- \pi^{(i+1)}}\left(\delta_{nd}^{(i)}\right)^2.
 \end{align}
The iterative estimation of $\gamma$ as just proposed has an advantage over using a constant value based on the volume of the data, as done in robust mixture models \cite{coretto2016robust}. Indeed, $\gamma$ is updated using the actual volume occupied by the outliers, which increases the ability of the algorithm to discriminate between inliers and outliers.

%

Another prominent advantage of DeepGUM for robustly predicting multidimensional outputs is its flexibility for 
handling the granularity of outliers. Consider for example to problem of locating landmarks in an image. One may want to devise a method that disregards outlying landmarks and not the whole image. In this case, one may use a GUM model for each landmark category. In the case of two-dimensional landmarks, this induces $D/2$ covariance matrices of size 2 ($D$ is the dimensionality of the target space). Similarly one may use an coordinate-wise outlier model, namely $D$ scalar variances. Finally, one may use an image-wise outlier model, i.e. the model detailed above. This flexibility is an attractive property of the proposed model as opposed to \cite{Belagiannis2015} which uses a coordinate-wise outlier model.

%

\subsection{Network Loss Function}
\label{sec:bp}
As already mentioned we use SGD to estimate the network parameters $\wvect$. Given the updated GUM parameters estimated with EM, $\thetavect^{(c)}$, the regression loss function is weighted with the responsibility of each data pair:
\begin{align}
\label{eq:loss}
\mathcal{L}_{\textsc{deepgum}}=\sum_{n=1}^Nr_{n}(\thetavect^{(c)}){||\lown-\phivect (\highn ; \wvect)||}^2_2.
\end{align}
\begin{figure}[t!]
 \centering
 \includegraphics[width=\linewidth]{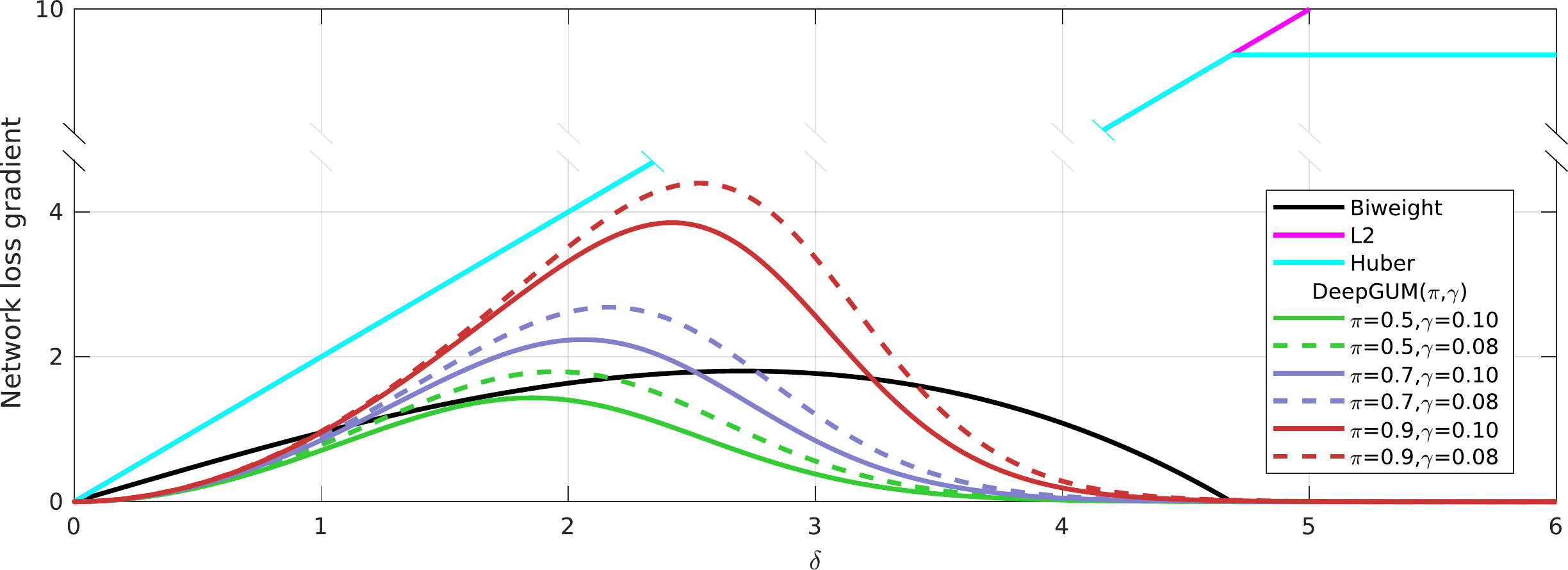} 
 \caption{Loss gradients for Biweight (black), Huber (cyan), $L_2$ (magenta), and DeepGUM (remaining colors).
 Huber and $L_2$ overlap up to $\delta=4.6851$ (the plots are truncated along the vertical coordinate). DeepGUM is shown for different values of $\pi$ and $\gamma$, although in practice they are estimated via EM. The gradients of DeepGUM and Biweight vanish for large residuals. DeepGUM offers some flexibility over Biweight thanks to $\pi$ and $\gamma$.\label{fig:losses}}
\end{figure}
With this formulation, the contribution of a training pair to the loss gradient vanishes (i) if the sample is an inlier with small error  ($\|\deltavect_n\|_2\rightarrow0,r_n\rightarrow1$) or (ii) if the sample is an outlier ($r_n\rightarrow0$). In both cases, the network will not back propagate any error. Consequently, the parameters $\wvect$ are updated only with inliers. This is graphically shown in Figure~\ref{fig:losses}, where we plot the loss gradient as a function of a one-dimensional residual $\delta$, for DeepGUM, Biweight, Huber and $L_2$. For fair comparison with Biweight and Huber, the plots correspond to a unit variance (i.e. standard normal, see discussion following eq.~(3) in~\cite{Belagiannis2015}). We plot the DeepGUM loss gradient for different values of $\pi$ and $\gamma$ to discuss different situations, although in practice all the parameters are estimated with EM. We observe that the gradient of the Huber loss increases linearly with $\delta$, until reaching a stable point (corresponding to $c=4.6851$ in~\cite{Belagiannis2015}). Conversely, the gradient of both DeepGUM and Biweight vanishes for large residuals (i.e. $\delta>c$). Importantly, DeepGUM offers some flexibility as compared to Biweight. Indeed, we observe that when the amount of inliers increases (large $\pi$) or the spread of outliers increases (small $\gamma$), the importance given to inliers is higher, which is a desirable property. The opposite effect takes place for lower amounts of inliers and/or reduced outlier spread.
%


%

\subsection{Training Algorithm}
In order to train the proposed model, we assume the existence of a training and validation datasets, denoted 
$\mathcal{T}=\{\high_n^\textsc{t},\low_n^\textsc{t}\}_{n=1}^{N_\textsc{t}}$ and
$\mathcal{V}=\{\high_n^\textsc{v},\low_n^\textsc{v}\}_{n=1}^{N_\textsc{v}}$, respectively. 
The training alternates between the unsupervised EM algorithm of Section~\ref{sec:em} 
and the supervised SGD algorithm of Section~\ref{sec:bp}, i.e. Algorithm~\ref{alg:DeepGUM}. EM takes as input the training set, alternates between responsibility evaluation, \eqref{eq:rn} and mixture parameter update, \eqref{eq:sigmamat}, \eqref{eq:pi}, \eqref{eq:lambda}, and iterates until convergence, namely until the mixture parameters do not evolve anymore. The current mixture parameters are used to evaluate the responsibilities of the validation set. The SGD algorithm takes as input the training and validation sets as well as the associated responsibilities. In order to prevent over-fitting, we perform early stopping on the validation set with a patience of $K$ epochs.

Notice that the training procedure requires neither specific annotation of outliers nor the ratio of outliers present in the data. The procedure is initialized by executing SGD, as just described, with all the samples being supposed to be inliers, i.e. $r_n=1, \forall n$. Algorithm~\ref{alg:DeepGUM} is stopped when $\mathcal{L}_\textsc{DEEPGUM}$ does not decrease anymore.
It is important to notice that we do not need to constrain the model to avoid the trivial solution, namely all the samples are considered as outliers. This is because after the first SGD execution, the network can discriminate between the two categories. In the extreme case when DeepGUM would consider all the samples as outliers, the algorithm would stop after the first SGD run and would output the initial model.

\begin{algorithm}[t]
   \caption{DeepGUM training}
   \label{alg:DeepGUM}
\begin{algorithmic}
   \STATE {\bfseries input:} $\mathcal{T}=(\high_n^\textsc{t},\low_n^\textsc{t})_{n=1}^{N_\textsc{t}}$, $\mathcal{V}=\{\high_n^\textsc{v},\low_n^\textsc{v}\}_{n=1}^{N_\textsc{v}}$,  and $\epsilon>0$ (convergence threshold).
   \STATE {\bfseries initialization: } Run SGD on $\mathcal{T}$ to minimize~(\ref{eq:loss}) 
with $r_n=1, \forall n$, until the convergence criterion on $\mathcal{V}$ is reached.
  \REPEAT
    \STATE {\bfseries EM algorithm: } Unsupervised outlier detection
    \REPEAT
      \STATE Update the $r_n$'s with~(\ref{eq:rn}).
      \STATE Update the mixture parameters with~(\ref{eq:sigmamat}),~(\ref{eq:pi}),~(\ref{eq:lambda}).
    \UNTIL{The parameters $\thetavect$ are stable.}
    \STATE{\bfseries SGD: } Deep regression learning
    \REPEAT
      \STATE Run SGD to minimize $\mathcal{L}_\textsc{DEEPGUM}$ in~(\ref{eq:loss}).
    \UNTIL{Early stop with a patience of $K$ epochs.}
 \UNTIL{$\mathcal{L}_\textsc{DEEPGUM}$ grows on $\mathcal{V}$.}
\end{algorithmic}
\end{algorithm}
Since EM provides the data covariance matrix $\Sigmamat$, it may be tempting to use the Mahalanobis norm instead of the $L_2$ norm in (\ref{eq:loss}). The covariance matrix is narrow along output dimensions with low-amplitude noise and wide along dimensions with high-amplitude noise. The Mahalanobis distance would give equal importance to low- and high-amplitude noise dimensions which is not desired. 
Another interesting feature of the proposed algorithm is that the posterior $r_n$ weights the learning rate of sample $n$ as its gradient is simply multiplied by $r_n$. Therefore, the proposed algorithm automatically selects a learning rate for each individual training sample.

\section{Experiments}
\label{expe}
The purpose of the experimental validation is two-fold. First, we empirically validate DeepGUM with three datasets that are naturally corrupted with outliers. The validations are carried out with the following applications: fashion landmark detection (Section \ref{fld}), age estimation (Section \ref{rae}) and head pose estimation (Section \ref{head}). Second, we delve into the robustness of DeepGUM and analyze its behavior in comparison with existing robust deep regression techniques by corrupting the annotations with an increasing percentage of outliers on the facial landmark detection task (Section~\ref{fald}).

We systematically compare DeepGUM with the standard $L_2$ loss, the Huber loss and the Biweight loss (used in~\cite{Belagiannis2015}). In all these cases, we use the VGG-16 architecture~\cite{Simonyan2014} pre-trained on ImageNet~\cite{ILSVRC15}. We also tried to use the architecture proposed in~\cite{Belagiannis2015}, but we were unable to reproduce the results reported in \cite{Belagiannis2015} on the LSP and Parse datasets, using the code provided by the authors. Therefore, for the sake of reproducibility and for a fair comparison between different robust loss functions, we used VGG-16 in all our experiments. Following the recommendations from \cite{lathuiliere2018comprehensive}, we fine-tune the last convolutional block and both fully connected layers with a mini-batch of size 128 and learning rate set to $10^{-4}$. The fine-tuning starts with 3 epochs of $L_2$ loss, before exploiting either the Biweight, Huber of DeepGUM loss. When using any of these three losses, the network output is normalized with the median absolute deviation (as in~\cite{Belagiannis2015}), computed on the entire dataset after each epoch. Early stopping with a patience of $K=5$ epochs is employed and the data is augmented using mirroring.

In order to evaluate the methods, we report the mean absolute error (MAE) between the regression target and the network output over the test set. Inspired by \cite{lathuiliere2018comprehensive}, we complete the evaluation with statistical tests that allow to point out when the differences between methods are systematic and statistically significant or due to chance. Statistical tests are run per-image regression errors and therefore can only be applied to the methods for which the code is available, and not to average errors reported in the literature; in the latter case, only MAE are made available. In practice, we use the non-parametric Wilcoxon signed-rank test~\cite{wilcoxon:test} to assess whether the null hypothesis (the median difference between pairs of observations is zero) is true or false. We denote the statistical significance with $^*$, $^{**}$ or $^{***}$, corresponding to a $p$-value (the conditional probability of, given the null hypothesis is true, getting a test statistic as extreme or more extreme than the calculated test statistic) smaller than $p=0.05$, $p=0.01$ or $p=0.001$, respectively. We only report the statistical significance of the methods with the lowest MAE. For instance, A$^{***}$ means that \textit{the probability that method} \textrm{A} \textit{is equivalent to any other method is less than $p=0.001$}.


\subsection{Fashion Landmark Detection} 
\label{fld}

Visual  fashion  analysis  presents a  wide spectrum of applications such as cloth recognition, retrieval, and recommendation. We employ the fashion landmark dataset (FLD)~\cite{liu2016deepfashion} that includes more than $120K$ images, where each image is labeled with eight landmarks. The dataset is equally divided in three subsets: upper-body clothes (6 landmarks), full-body clothes (8 landmarks) and lower-body clothes (4 landmarks). We randomly split each subset of the dataset into test ($5K$), validation ($5K$) and training (${\sim}30K$). 
Two metrics are used: the mean absolute error (MAE) of the landmark localization and the percentage of failures (landmarks detected further from the ground truth than a given threshold). We employ \emph{landmark-wise} 
$r_{n}$. 



\begin{table}[t]
\centering
 \caption{Mean absolute error on the upper-body subset of FLD, per landmark and in average. The 
landmarks are left (L) and right (R) collar (C), sleeve (S) and hem (H). The results of DFA are from~\cite{liu16FashionLandmark} and therefore do not take part in the statistical comparison.\label{tab:results-fld}\vspace{1mm}}
 \begin{tabular}{llllllll}
 \toprule
 \multirow{2}{*}{Method} & \multicolumn{7}{c}{Upper-body landmarks} \\
 \cmidrule{2-8}
  & LC & RC & LS & RS & LH & RH & Avg. \\
  \midrule
  DFA~\cite{liu16FashionLandmark} ($L_2$) & 15.90 & 15.90 & 30.02 & 29.12 & 23.07 & 22.85 & 22.85 \\
  DFA~\cite{liu16FashionLandmark} (5 VGG) & 10.75 & 10.75 & 20.38 & 19.93 & 15.90 & 16.12 & 15.23\\
 \midrule
  $L_2$ & 12.08 & 12.08 & 18.87 & 18.91 & 16.47 & 16.40 & 15.80 \\
  Huber~\cite{huber1964robust} & 14.32 & 13.71 & 20.85 & 19.57 & 20.06 & 19.99 & 18.08 \\
  Biweight~\cite{Belagiannis2015} & 13.32 & 13.29& 21.88 &21.84 &18.49 & 18.44 & 17.88 \\
  DeepGUM & 11.97$^{***}$ & 11.99$^{***}$ & 18.59$^{***}$ & 18.50$^{***}$ & 16.44$^{***}$ & 16.29$^{***}$ & 15.63$^{***}$ \\
  \bottomrule
 \end{tabular}
%
 
\end{table}


Table~\ref{tab:results-fld} reports the results obtained on the upper-body subset of the fashion landmark dataset (additional results on full-body and lower-body subsets are included in the supplementary material). We report the mean average error (in pixels) for each landmark individually, and the overall average (last column). While for the first subset we can compare with the very recent results reported in~\cite{liu16FashionLandmark}, for the other there are no previously reported results. Generally speaking, we outperform all other baselines in average, but also in each of the individual landmarks. The only exception is the comparison against the method utilizing five VGG pipelines to estimate the position of the landmarks. Although this method reports slightly better performance than DeepGUM for some columns of Table~\ref{tab:results-fld}, we recall that we are using one single VGG as front-end, and therefore the representation power cannot be the same as the one associated to a pipeline employing five VGG's trained for tasks such as pose estimation and cloth classification that clearly aid the fashion landmark estimation task.

\begin{figure}[t]
  \captionsetup[subfigure]{labelformat=empty}
  \centering
  \subfloat[]{\includegraphics[height=0.23\columnwidth]{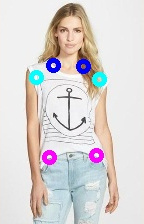}}
  \subfloat[]{\includegraphics[height=0.23\columnwidth]{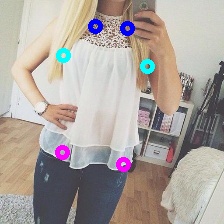}}
  \subfloat[]{\includegraphics[height=0.23\columnwidth]{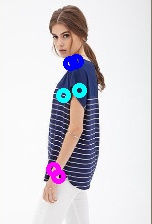}}    
  \subfloat[]{\includegraphics[height=0.23\columnwidth]{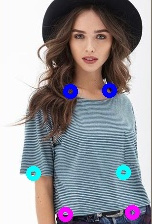}}
  \subfloat[]{\includegraphics[height=0.23\columnwidth]{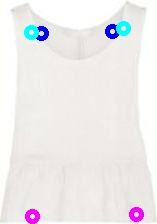}}
  \subfloat[]{\includegraphics[height=0.23\columnwidth]{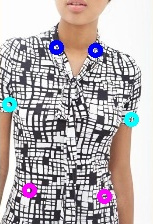}}\vspace{-5mm}
  \caption{Sample fashion landmarks detected by DeepGUM.\vspace{-4mm}}
\label{fig:fld-examples}
\end{figure}

Interestingly, DeepGUM yields better results than $L_2$ regression and a major improvement over Biweight~\cite{Belagiannis2015} and Huber~\cite{huber1964robust}. This behavior is systematic for all fashion landmarks and statistically significant (with $p<0.001$). In order to better understand this behavior, we computed the percentage of outliers detected by DeepGUM and Biweight, which are $3\%$ and $10\%$ respectively (after convergence). We believe that within this difference ($7\%$ corresponds to $2.1K$ images) there are mostly ``difficult'' inliers, from which the network could learn a lot (and does it in DeepGUM) if they were not discarded as happens with Biweight. This illustrates the importance of rejecting the outliers while keeping the inliers in the learning loop, and exhibits the robustness of DeepGUM in doing so. Figure~\ref{fig:fld-examples} displays a few landmarks estimated by DeepGUM.


\subsection{Age Estimation}
\label{rae}
Age estimation from a single face image is an important task in computer vision with applications in access control and human-computer interaction. This task is closely related to the prediction of other biometric  and  facial  attributes, such  as  gender, ethnicity,  and hair  color.  We use the cross-age celebrity dataset (CACD)~\cite{Chen2014b} that contains $163,446$ images from $2,000$ celebrities. The images are collected from search engines using the celebrity's name and desired year (from 2004 to 2013).  The  dataset  splits  into 3  parts,  $1,800$  celebrities  are  used  for  training,  $80$  for validation and $120$ for testing. The validation and test sets are manually cleaned whereas the training set is noisy.  In our experiments, we report results using \emph{image-wise} $r_{n}$.

Apart from DeepGUM, $L_2$, Biweight and Huber, we also compare to the age estimation method based on deep expectation (Dex)~\cite{rothe2016deep}, which was the winner of the Looking at People 2015 challenge. This method uses the VGG-16 architecture and poses the age estimation problem as a 
classification problem followed by a softmax expected value refinement. Regression-by-classification strategies have also been proposed for memorability and virality~\cite{siarohin2017make,alameda2017viraliency}. We report results with two different approaches using Dex. First, our implementation of the original Dex model. Second, we add the GUM model on top the the Dex architecture; we termed this architecture DexGUM. 

 
%

\begin{figure}[t!]
\begin{minipage}[l]{0.25\textwidth}
   \begin{tabular}{ll}
  	\toprule
        Method & MAE \\
  	\midrule
        $L_2$ &  $5.75$   \\
        Huber~\cite{huber1964robust} & 5.59 \\
        Biweight~\cite{Belagiannis2015} & $5.55$\\
        Dex~\cite{rothe2016deep}&  $5.25$   \\
        DexGUM$^{***}$ &  $5.14$   \\
        DeepGUM$^{***}$ &  $5.08$   \\
        \toprule
  \end{tabular}
\end{minipage}
\begin{minipage}[r]{0.75\textwidth}
  \captionsetup[subfigure]{labelformat=empty}
  \centering
  \subfloat[14]{\includegraphics[width=0.16\columnwidth]{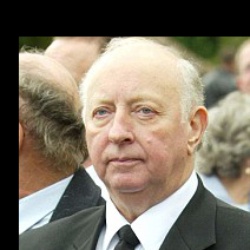}}
  \subfloat[14]{\includegraphics[width=0.16\columnwidth]{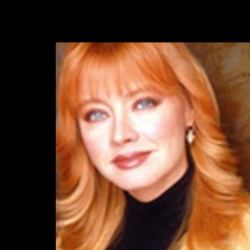}}
  \subfloat[14]{\includegraphics[width=0.16\columnwidth]{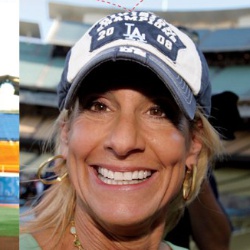}}    
  \subfloat[16]{\includegraphics[width=0.16\columnwidth]{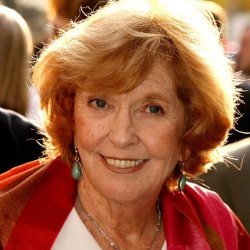}}
  \subfloat[20]{\includegraphics[width=0.16\columnwidth]{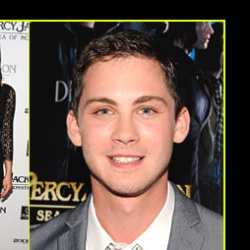}}
  \subfloat[23]{\includegraphics[width=0.16\columnwidth]{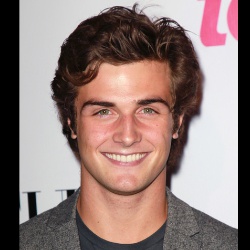}}\\\vspace{-0.3cm}%
  \subfloat[49]{\includegraphics[width=0.16\columnwidth]{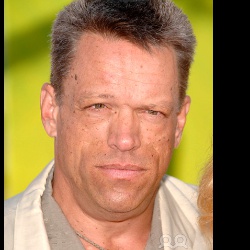}}
  \subfloat[51]{\includegraphics[width=0.16\columnwidth]{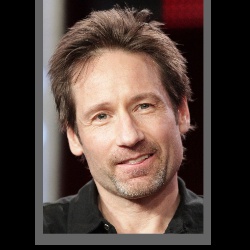}}
  \subfloat[60]{\includegraphics[width=0.16\columnwidth]{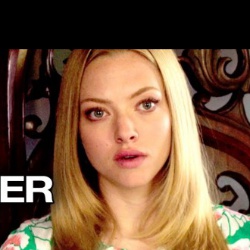}}    
  \subfloat[60]{\includegraphics[width=0.16\columnwidth]{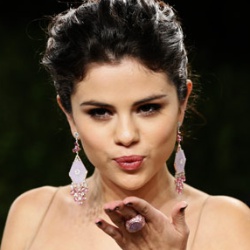}}
  \subfloat[60]{\includegraphics[width=0.16\columnwidth]{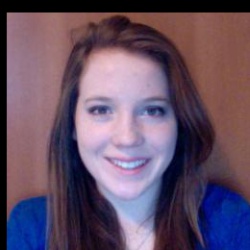}}
  \subfloat[62]{\includegraphics[width=0.16\columnwidth]{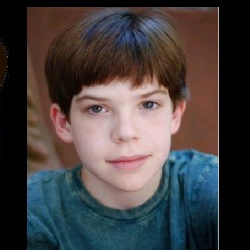}}
  \vspace{-0.3cm}
\end{minipage}
\caption{Results on the CACD dataset: (left) mean absolute error and (right) images considered as outliers by DeepGUM, the annotation is displayed below each 
image.\label{fig:results-cacd}\vspace{-5mm}}
\end{figure}

The table in Figure~\ref{fig:results-cacd} reports the results obtained on the CACD test set for age estimation. We report the mean absolute error (in years) for size different methods. We can easily observe that DeepGUM exhibits the best results: $5.08$ years of MAE ($0.7$ years better than $L_2$). Importantly, the architectures using GUM (DeepGUM followed by DexGUM) are the ones offering the best performance. This claim is supported by the results of the statistical tests, which say that DexGUM and DeepGUM are statistically better than the rest (with $p<0.001$), and that there are no statistical differences between them. This is further supported by the histogram of the error included in the supplementary material. DeepGUM considered that $7\%$ of images were outliers and thus these images were undervalued during training. The images in Figure~\ref{fig:results-cacd} correspond to outliers detected by DeepGUM during training, and illustrate the ability of DeepGUM to detect outliers. Since the dataset was automatically annotated, it is prone to corrupted annotations. Indeed,  the age of each celebrity is automatically annotated by subtracting the date of birth from the picture time-stamp. Intuitively, this procedure is problematic since it assumes that the automatically collected and annotated images show the right celebrity and that the times-tamp and date of birth are correct. Our experimental evaluation clearly demonstrates the benefit of a robust regression technique to operate on datasets populated with outliers.


\subsection{Head Pose Estimation}
\label{head}
The McGill real-world face video dataset \cite{Demirkus:2015} consists of 60 videos (a single participant per video, 31 women
and 29 men) recorded with the goal of studying unconstrained face classification. The videos were recorded in both indoor and outdoor environments under different illumination conditions and participants move freely. Consequently, some frames suffer from important occlusions. The yaw angle (ranging from $-90^\circ$ to $90^\circ$) is annotated using a two-step labeling procedure that, first, automatically provides the most probable angle as well as a degree of confidence, and then the final label is chosen by a human annotator among the plausible angle values. Since the resulting annotations are not perfect it makes this dataset suitable to benchmark robust regression models. As the training and test sets are not separated in the original dataset, we perform a 7-fold cross-validation. We report the fold-wise MAE average and standard deviation as well as the statistical significance corresponding to the concatenation of the test results of the 7 folds. Importantly, only a subset of the dataset is publicly available (35 videos over 60). 

In Table \ref{table:mcgill}, we report the results obtained with different methods and employ a dagger to indicate when a particular method uses the entire dataset (60 videos) for training. We can easily notice that DeepGUM exhibits the best results compared to the other ConvNets methods (respectively $0.99^\circ$, $0.50^\circ$ and $0.20^\circ$ lower than $L_2$, Huber and Biweight in MAE). The last three approaches, all using deep architectures, significantly outperform the current state-of-the-art approach~\cite{Drouard2017RobustHE}. Among them, DeepGUM is significantly better than the rest with $p<0.001$.

\begin{table}[t]
  \caption{Mean average error on the McGill dataset. The results of the first half of the table are directly taken from the respective papers and therefore no statistical comparison is possible. $^{\dagger}$Uses extra training data. 
    \label{table:mcgill}
  }
  \centering
  \begin{tabular}{lcc}
  	\toprule
        Method & MAE &RMSE\\
  	\midrule
        Xiong et al.~\cite{xiong13}$^{\dagger}$ &-& $29.81 \pm 7.73$\\
        
        Zhu and Ramanan~\cite{zhu2012face}$^{\dagger}$&-& $35.70 \pm 7.48$\\
        
        Demirkus et al.~\cite{Demirkus:2015}$^{\dagger}$ &-&$ 12.41\pm 1.60$\\
        
        Drouard et al.~\cite{Drouard2017RobustHE}
         & $ 12.22\pm6.42$ & $ 23.00\pm9.42$\\
         \midrule
         
        $L_2$ &  $8.60\pm1.18$  &$12.03\pm1.66 $\\
        
        Huber~\cite{huber1964robust} & $8.11\pm1.08$ & $11.79\pm1.59$ \\
        
        Biweight~\cite{Belagiannis2015} & $7.81\pm1.31$ & $11.56\pm1.95$\\
        
        DeepGUM$^{***}$ &  $ 7.61\pm1.00$  &$11.37\pm1.34$ \\
        \toprule
  \end{tabular}
  \vspace{-4mm}
\end{table}

\subsection{Facial Landmark Detection}
\label{fald}

We perform experiments on the LFW and NET facial landmark detection 
datasets~\cite{Sun2013} that consist of 5590 and 7876 face images, respectively. We combined both datasets and employed the same data partition as in~\cite{Sun2013}. Each face is labeled with the positions of five key-points in Cartesian coordinates, 
namely left and right eye, nose, and left and right corners of the mouth. The detection error is measured with the 
Euclidean distance between the estimated and the ground truth position of the landmark, divided by the width of the 
face image, as in~\cite{Sun2013}. The performance is measured with the failure rate of each landmark, where errors 
larger than $5\%$ are counted as failures. The two aforementioned datasets can be considered as outlier-free since the 
average failure rate reported in the literature falls below $1\%$. Therefore, we artificially modify the annotations 
of the datasets for facial landmark detection to find the breakdown point of DeepGUM. Our purpose is to study the 
robustness of the proposed deep mixture model to outliers generated in controlled conditions. We 
use three different types of outliers:
\begin{itemize}
 \item Normally Generated Outliers (\emph{NGO}): A percentage of landmarks is selected, regardless of whether 
they belong to the same image or not, and shifted a distance of $d$ pixels in a uniformly chosen random direction. The 
distance $d$ follows a Gaussian distribution, $\mathcal{N}(25, 2)$. \emph{NGO} simulates errors produced 
by human annotators that made a mistake when clicking, thus annotating in a slightly wrong location.
 \item Local - Uniformly Generated Outliers (\emph{l-UGO}): It follows the same philosophy as \emph{NGO}, 
sampling the distance $d$ from a uniform distribution over the image, instead of a Gaussian. Such errors simulate human 
errors that are not related to the human precision, such as not selecting the point or misunderstanding the image.
 \item Global - Uniformly Generated Outliers (\emph{g-UGO}): As in the previous case, the landmarks are corrupted with 
uniform noise. However, in \emph{g-UGO} the landmarks to be corrupted are grouped by image. In other words, we do not 
corrupt a subset of all landmarks regardless of the image they belong to, but rather corrupt all landmarks of a subset 
of the images. This strategy simulates problems with the annotation files or in the sensors in case of automatic 
annotation.
\end{itemize}
The first and the second types of outlier contamination employ \emph{landmark-wise} $r_{n}$, while the third uses \emph{image-wise} $r_{n}$. 

\begin{figure}[t]
  \centering
  \includegraphics[width=0.72\textwidth]{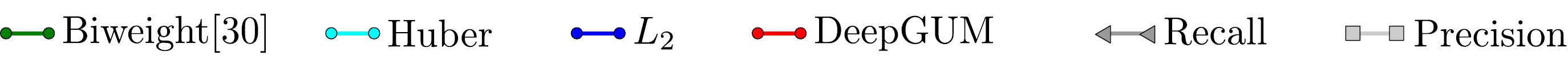}\vspace{-4mm}\\
  \subfloat{\includegraphics[width=0.3333\textwidth]{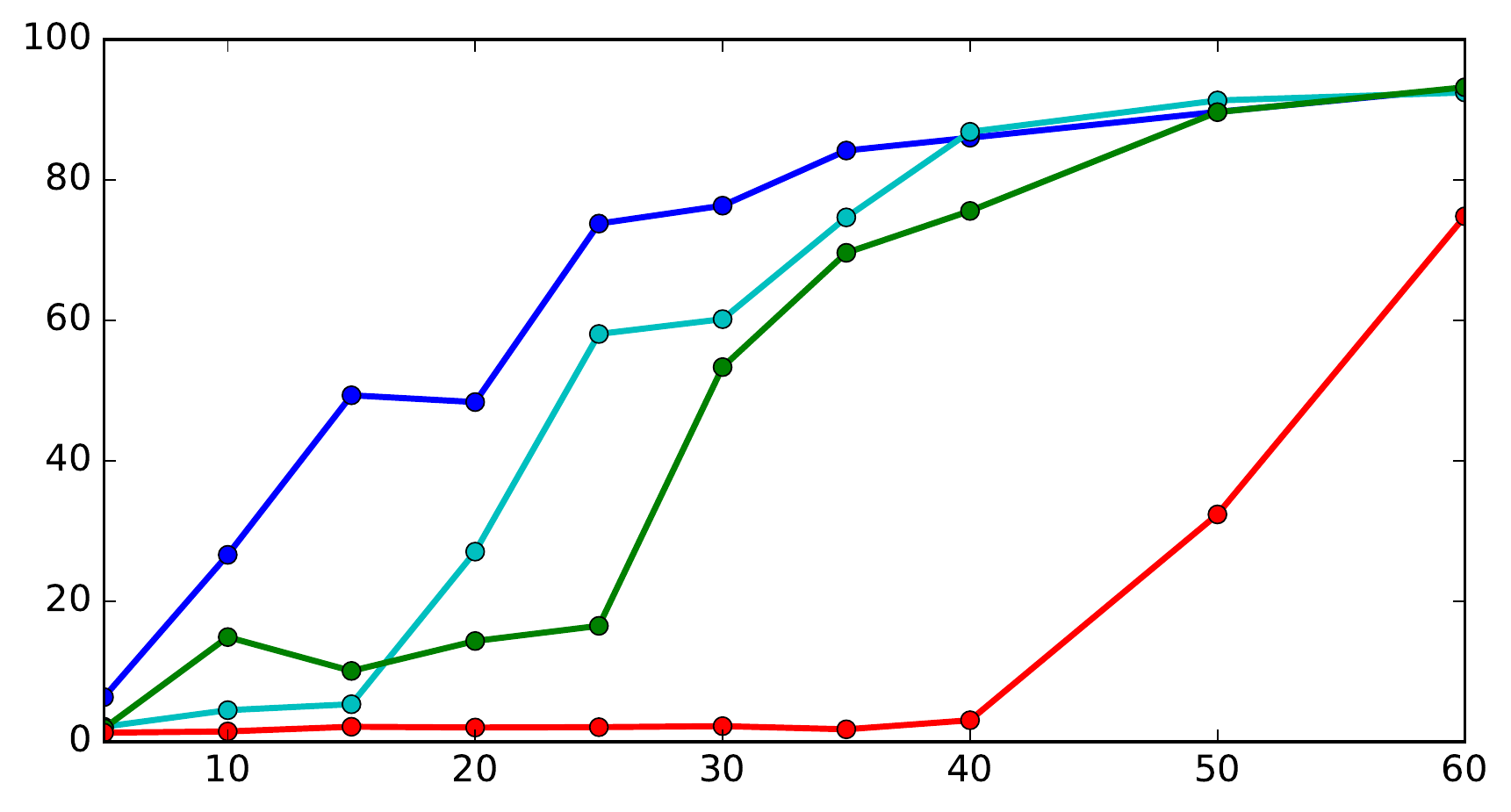}}
  \subfloat{\includegraphics[width=0.3333\textwidth]{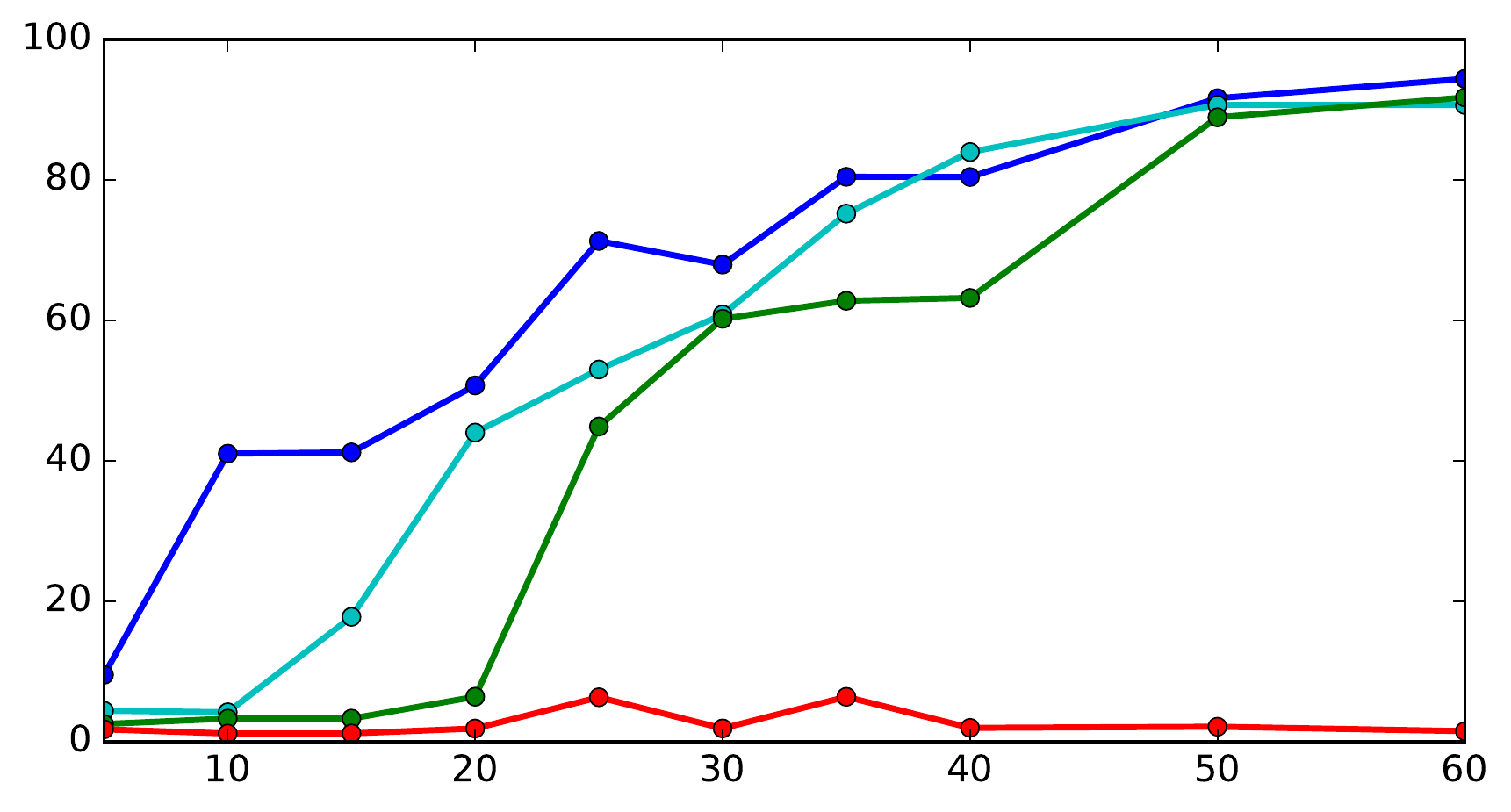}}
  \subfloat{\includegraphics[width=0.3333\textwidth]{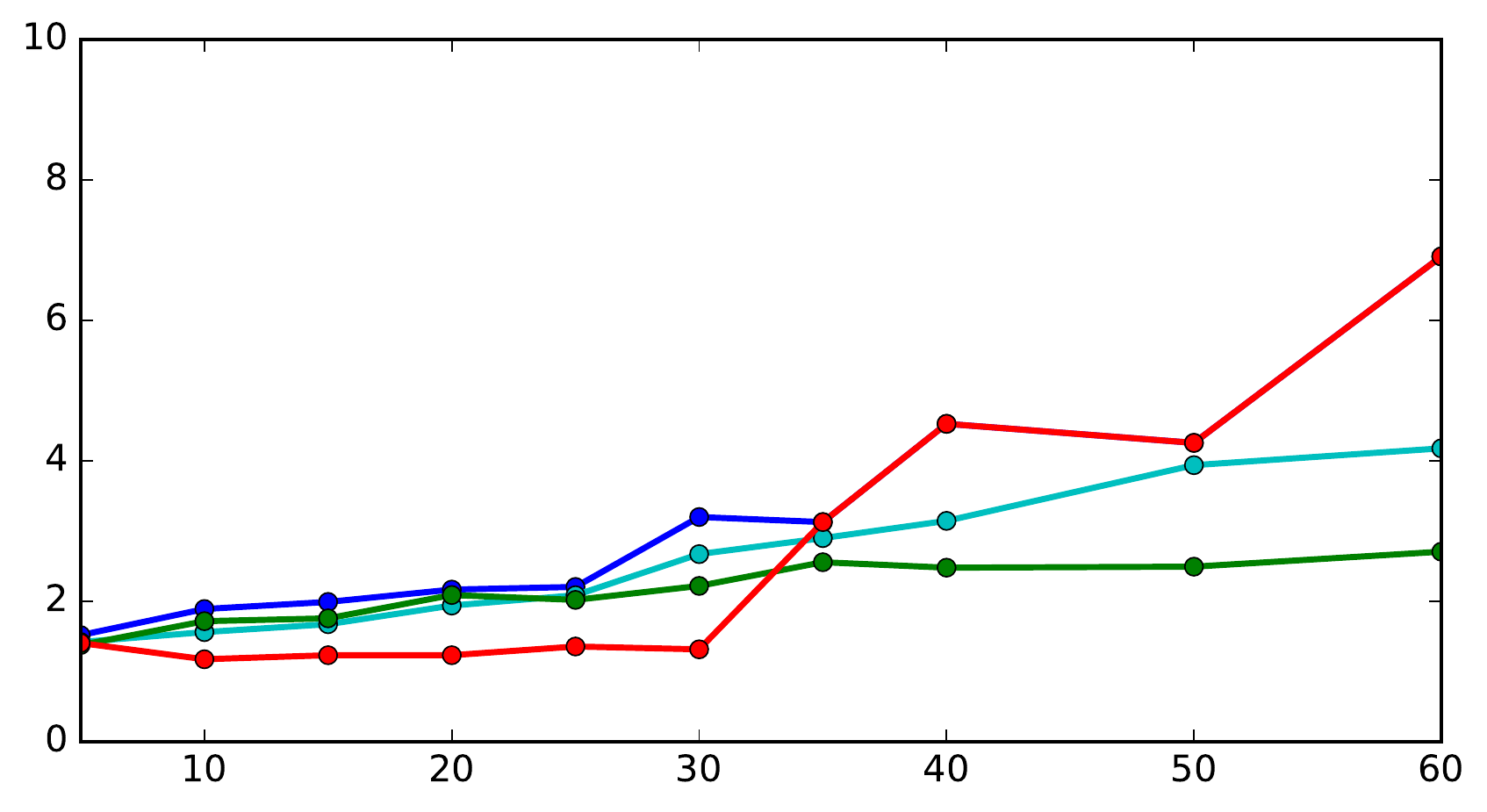}}\vspace{-3mm}\\
  \setcounter{subfigure}{0}
  \subfloat[\vspace{-2mm}\emph{l-UGO}]{\includegraphics[width=0.3333\textwidth]{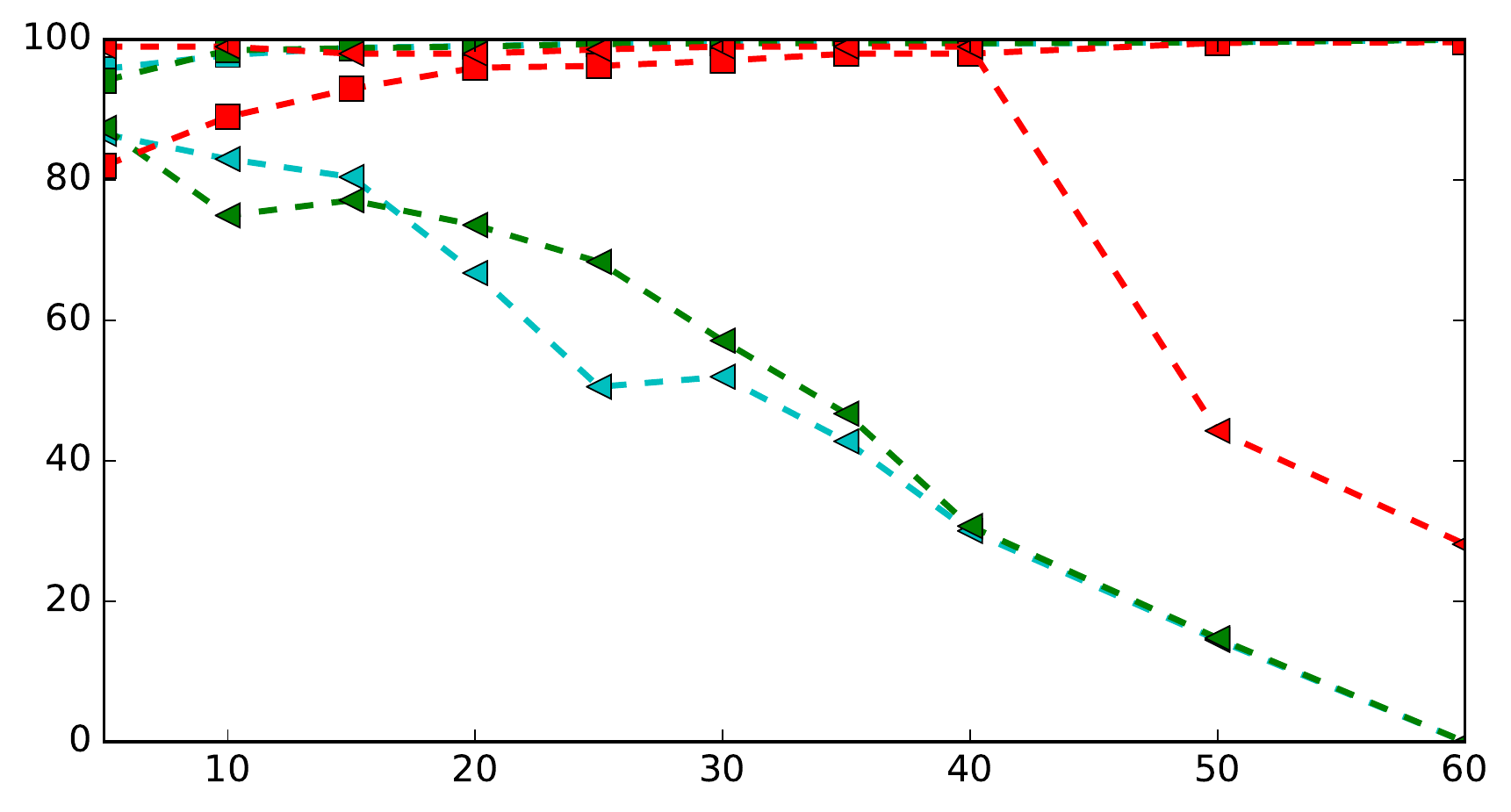}\label{6classchoiAprime}}
  \subfloat[\vspace{-2mm}\emph{g-UGO}]{\includegraphics[width=0.3333\textwidth]{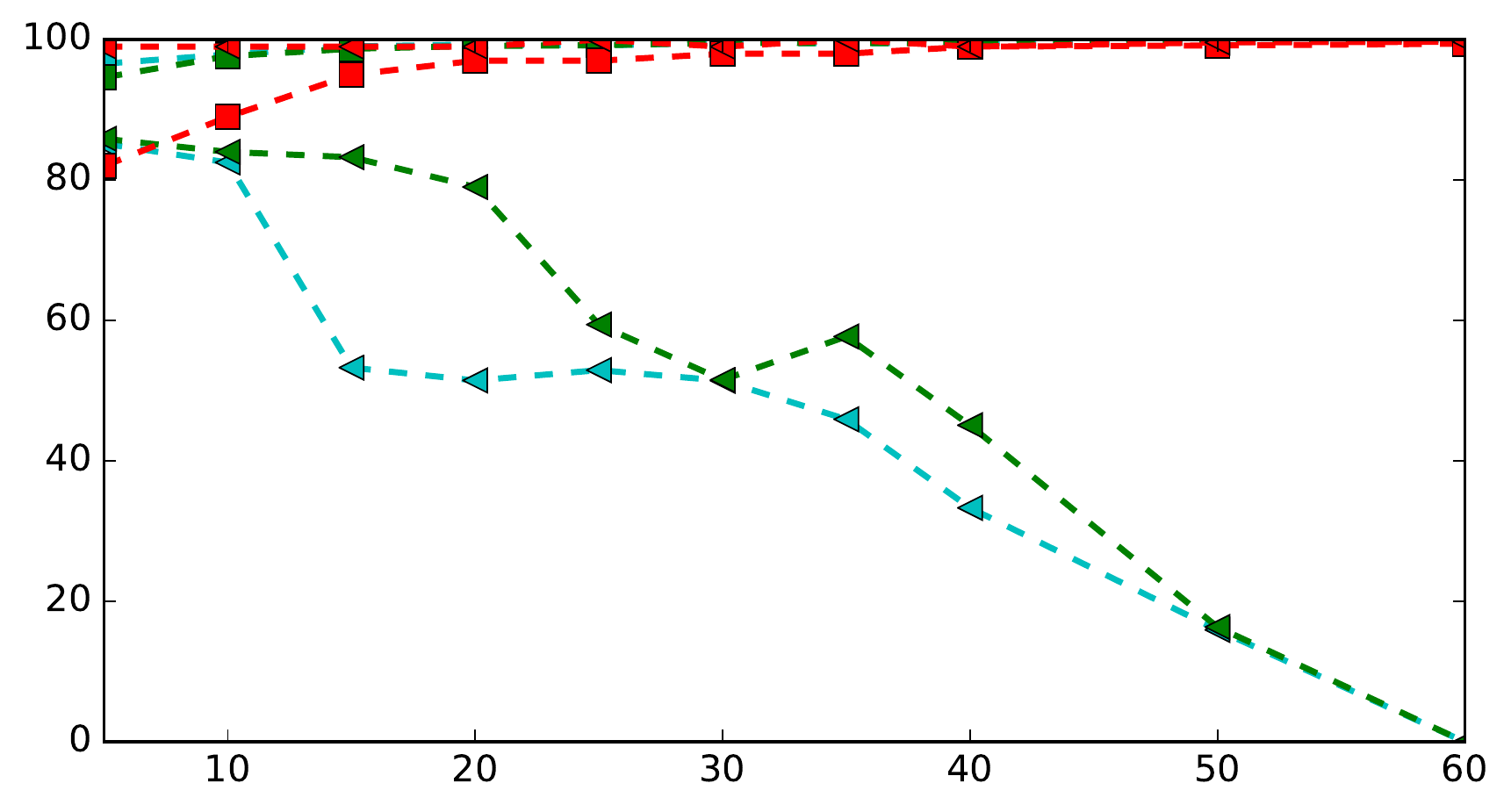}\label{6classchoiA}} 
  \subfloat[\vspace{-2mm}\emph{NGO}]{\includegraphics[width=0.3333\textwidth]{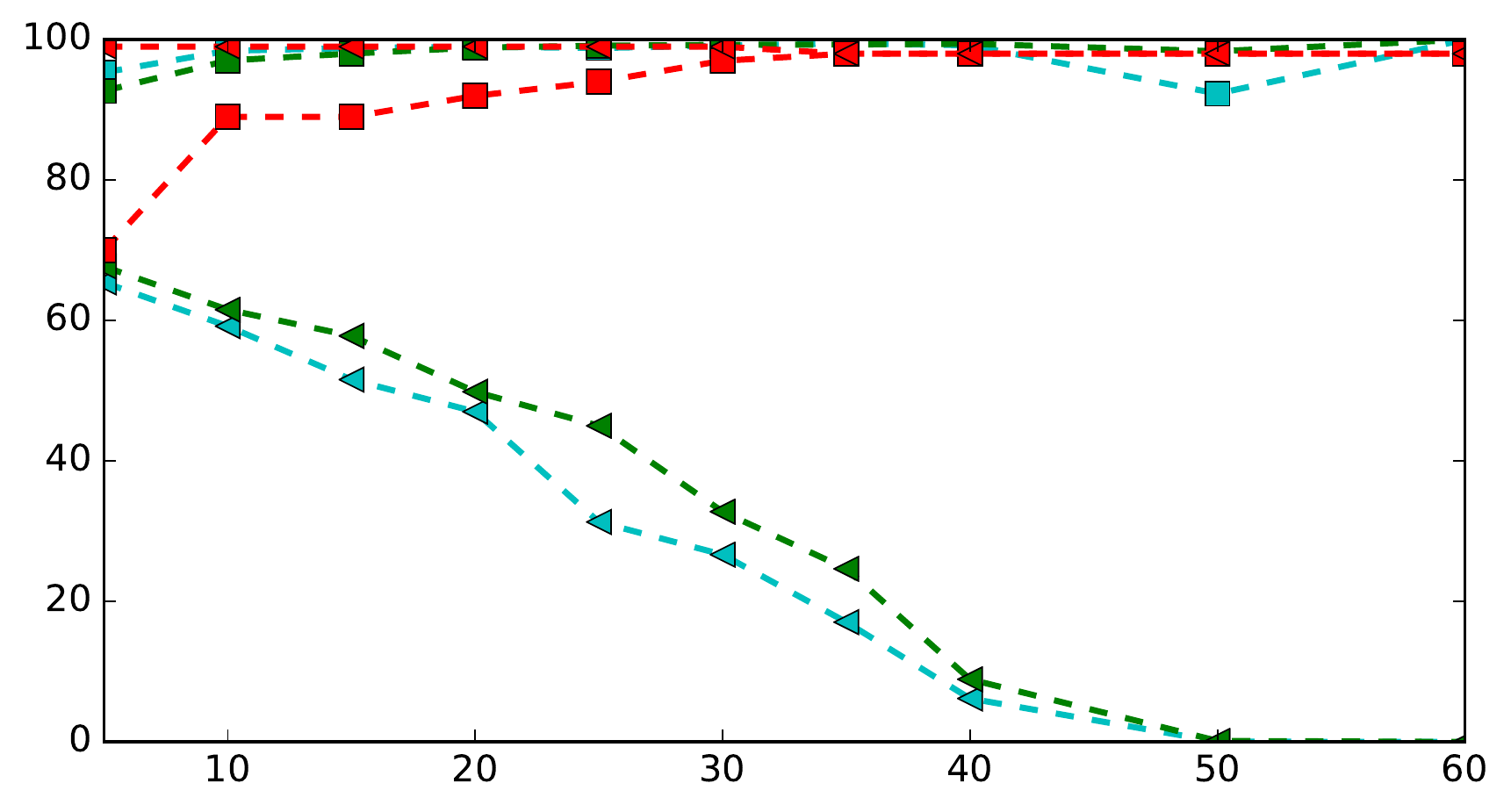}\label{5classSun}}
    \caption{Evolution of the failure rate (top)  when augmenting the noise for the 3 types of outliers considered. We also display the corresponding precisions and recalls in percentage (bottom) for the outlier class. Best seen in color.}\vspace{-5mm}
\label{fig:synthetic_outliers}
\end{figure}

The plots in Figure~\ref{fig:synthetic_outliers} report the failure rate of DeepGUM , Biweight, Huber and $L_2$ (top) on the clean test set and the outlier detection precision and recall of all except for $L_2$ (bottom) for the three types of synthetic noise on the corrupted training set. The precision corresponds to the percentage of training samples classified as outliers that are true outliers; and the recall corresponds to the percentage of outliers that are classified as such. The first conclusion that can be drawn directly from this figure are that, on the one hand, Biweight and Huber systematically present a lower recall than DeepGUM. In other words, DeepGUM exhibits the highest reliability at identifying and, therefore, ignoring outliers during training. And, on the other hand, DeepGUM tends to present a lower failure rate than Biweight, Huber and $L_2$ in most of the scenarios contemplated.

Regarding the four most-left plots, \emph{l-UGO} and \emph{g-UGO}, we can clearly observe that, while for limited amounts of outliers (i.e.\ $<10\%$) all methods report comparable performance, DeepGUM is clearly superior to $L_2$, Biweight and Huber for larger amounts of outliers. We can also safely identify a breakdown point of DeepGUM on \emph{l-UGO} at $\thicksim40\%$. This is inline with the reported precision and recall for the outlier detection task. While for Biweight and Huber, both decrease when increasing the number of outliers, these measures are constantly around $99\%$ for DeepGUM (before $40\%$ for \emph{l-UGO}). The fact that the breakdown point of DeepGUM under \emph{g-UGO} is higher than $50\%$ is due to fact that the a priori model of the outliers (i.e. uniform distribution) corresponds to the way the data is corrupted.

For \emph{NGO}, the corrupted annotation is always around the ground truth, leading to a failure rate smaller than 7$\%$ for all methods. We can see that all four methods exhibit comparable performance up to $30\%$ of outliers. Beyond that threshold, Biweight outperforms the other methods in spite of presenting a progressively lower recall and a high precision (i.e. Biweight identifies very few outliers, but the ones identified are true outliers). This behavior is also exhibited by Huber. Regarding DeepGUM, we observe that in this particular setting the results are aligned with $L_2$. This is because the SGD procedure is not able to find a better optimum after the first epoch and therefore the early stopping mechanism is triggered and SFD output the initial network, which corresponds to $L_2$. We can conclude that the strategy of DeepGUM, consisting in removing all points detected as outliers, is not effective in this particular experiment. In other words, having more noisy data is better than having only few clean data in this particular case of 0-mean highly correlated noise. Nevertheless, we consider an attractive property of DeepGUM the fact that it can automatically identify these particular cases and return an acceptable solution. 

\section{Conclusions}
\label{sec:conclusions}
This paper introduced a deep robust regression learning method that uses a Gaussian-uniform mixture model. The novelty of the paper resides in combining a probabilistic robust mixture model with deep learning in a jointly trainable fashion. In this context, previous studies only dealt with the classical $L_2$ loss function or Tukey's Biweight function, an M-estimator robust to outliers~\cite{Belagiannis2015}. Our proposal yields better 
performance than previous deep regression approaches by proposing a novel technique, and 
the derived optimization procedure, that alternates between the unsupervised task of outlier detection and the 
supervised task of learning network parameters. The experimental validation addresses four different tasks: 
facial and fashion landmark detection, age estimation, and head pose estimation. We have empirically shown 
that DeepGUM  (i) is a robust deep regression approach that does not need to rigidly specify \emph{a priori} the distribution 
(number and spread) of outliers, (ii) exhibits a higher breakdown point than existing methods when the 
outliers are sampled from a uniform distribution (being able to deal with more than $50\%$ of outlier contamination 
without providing incorrect results), and (iii) is capable of providing comparable or better results than current 
state-of-the-art approaches in the four aforementioned tasks. Finally, DeepGUM could be easily used to remove undesired samples that arise from tedious manual annotation. It could also deal with highly unusual training samples inherently present in automatically collected huge datasets, a problem that is currently addressed using  error-prone and time-consuming human supervision.

\noindent
{\textbf{Acknowledgments}.} This work was supported by the European Research Council via the ERC Advanced Grant VHIA (Vision and Hearing in Action) \#113340.

{\small
\bibliographystyle{splncs04}
\bibliography{referencesShort,extra}
}

\newpage
\title{Supplementary Material\\DeepGUM: Learning Deep Robust Regression with a Gaussian-Uniform Mixture Model} 
\titlerunning{DeepGUM: Learning Deep Robust Regression}

\authorrunning{S. Lathuili\`{e}re et al.}

%
\author{St\'{e}phane~Lathuili\`{e}re\inst{1,3},
  Pablo~Mesejo\inst{1,2},
  Xavier~Alameda-Pineda\inst{1}, 
  and~Radu~Horaud\inst{1}}
%

\authorrunning{S. Lathuili\`{e}re et al.}

\institute{Inria Grenoble Rh\^{o}ne-Alpes, Montbonnot-Saint-Martin, France, \and
  University of Granada, Granada, Spain,\and
  University of Trento, Trento, Italy\\
  \email{ {firstname.name}@inria.fr}}

\maketitle

\appendix
This document contains the supplementary material for the paper \textit{DeepGUM: Learning Deep Robust Regression with a Gaussian-Uniform Mixture Model}. We provide an extensive number of visual examples and extra-results obtained using our proposed probabilistic-based robust regression  
approach. The different sections of this document show additional information about the four tasks addressed in the paper.

\section{Fashion Landmark Detection}
\begin{figure*}[p]
  \captionsetup[subfigure]{labelformat=empty}
  \centering
\subfloat[]{\includegraphics[width=0.20\textwidth]{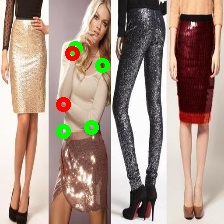}}
\subfloat[]{\includegraphics[width=0.20\textwidth]{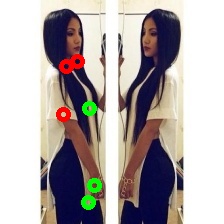}}
\subfloat[]{\includegraphics[width=0.20\textwidth]{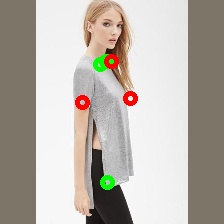}}
\subfloat[]{\includegraphics[width=0.20\textwidth]{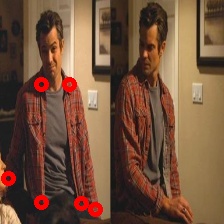}}
\subfloat[]{\includegraphics[width=0.20\textwidth]{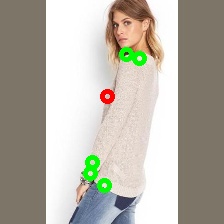}}\\
\subfloat[]{\includegraphics[width=0.20\textwidth]{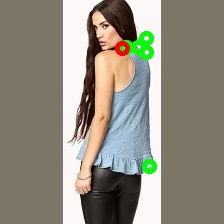}}
\subfloat[]{\includegraphics[width=0.20\textwidth]{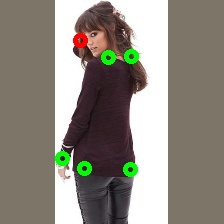}}
\subfloat[]{\includegraphics[width=0.20\textwidth]{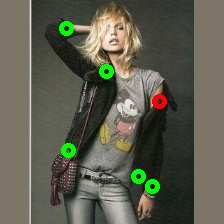}}
\subfloat[]{\includegraphics[width=0.20\textwidth]{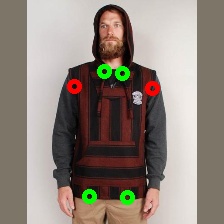}}
\subfloat[]{\includegraphics[width=0.20\textwidth]{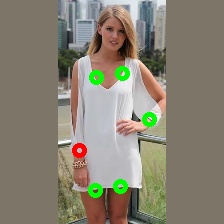}}\\
\subfloat[]{\includegraphics[width=0.20\textwidth]{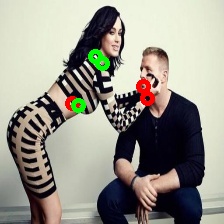}}
\subfloat[]{\includegraphics[width=0.20\textwidth]{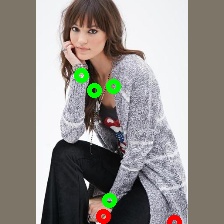}}
\subfloat[]{\includegraphics[width=0.20\textwidth]{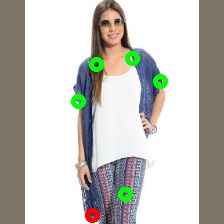}}
\subfloat[]{\includegraphics[width=0.20\textwidth]{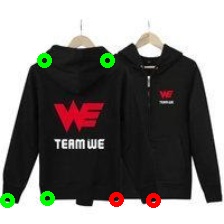}}
\subfloat[]{\includegraphics[width=0.20\textwidth]{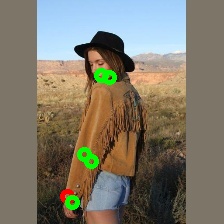}}\\
\subfloat[]{\includegraphics[width=0.20\textwidth]{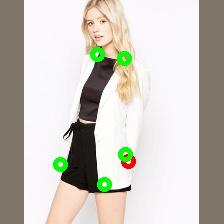}}
\subfloat[]{\includegraphics[width=0.20\textwidth]{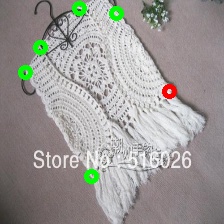}}
\subfloat[]{\includegraphics[width=0.20\textwidth]{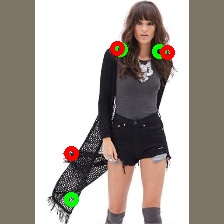}}
\subfloat[]{\includegraphics[width=0.20\textwidth]{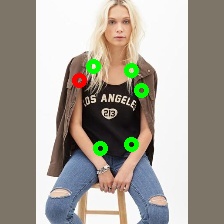}}
\subfloat[]{\includegraphics[width=0.20\textwidth]{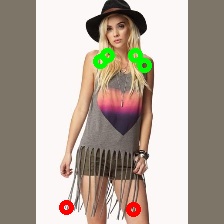}}\\
\subfloat[]{\includegraphics[width=0.20\textwidth]{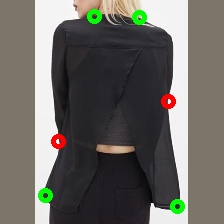}}
\subfloat[]{\includegraphics[width=0.20\textwidth]{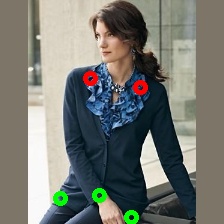}}
\subfloat[]{\includegraphics[width=0.20\textwidth]{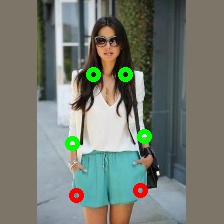}}
\subfloat[]{\includegraphics[width=0.20\textwidth]{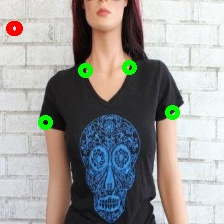}}
\subfloat[]{\includegraphics[width=0.20\textwidth]{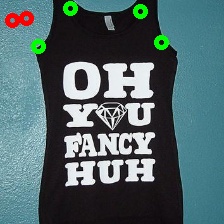}}    
      \caption{Example of images from the Fashion Landmark Dataset: landmarks detected as outliers by DeepGUM are shown 
in red, while inliers are shown in green.}
\label{fig:fashionOK}
\end{figure*}

\begin{figure*}[p]
  \captionsetup[subfigure]{labelformat=empty}
  \centering
\subfloat[]{\includegraphics[width=0.20\textwidth]{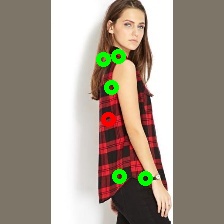}}
\subfloat[]{\includegraphics[width=0.20\textwidth]{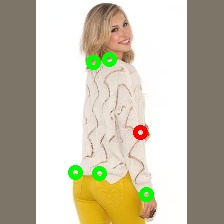}}
\subfloat[]{\includegraphics[width=0.20\textwidth]{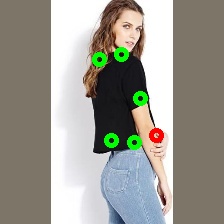}}
\subfloat[]{\includegraphics[width=0.20\textwidth]{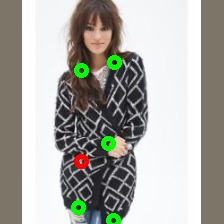}}
\subfloat[]{\includegraphics[width=0.20\textwidth]{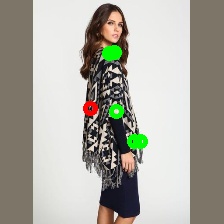}}\\
\subfloat[]{\includegraphics[width=0.20\textwidth]{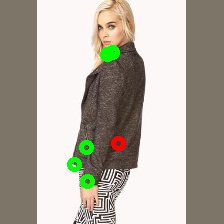}}
\subfloat[]{\includegraphics[width=0.20\textwidth]{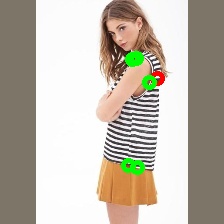}}
\subfloat[]{\includegraphics[width=0.20\textwidth]{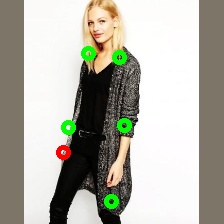}}
\subfloat[]{\includegraphics[width=0.20\textwidth]{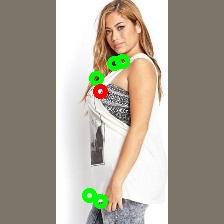}}
\subfloat[]{\includegraphics[width=0.20\textwidth]{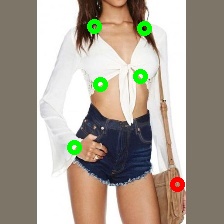}}\\
\subfloat[]{\includegraphics[width=0.20\textwidth]{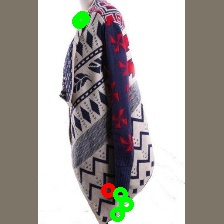}}
\subfloat[]{\includegraphics[width=0.20\textwidth]{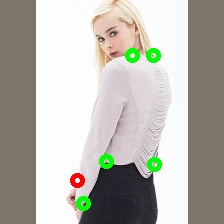}}
\subfloat[]{\includegraphics[width=0.20\textwidth]{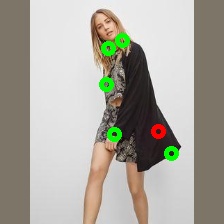}}
\subfloat[]{\includegraphics[width=0.20\textwidth]{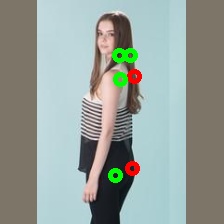}}
\subfloat[]{\includegraphics[width=0.20\textwidth]{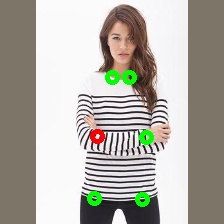}}\\
\subfloat[]{\includegraphics[width=0.20\textwidth]{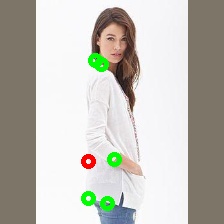}}
\subfloat[]{\includegraphics[width=0.20\textwidth]{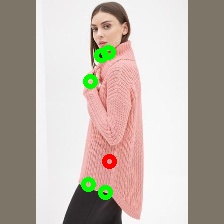}}
\subfloat[]{\includegraphics[width=0.20\textwidth]{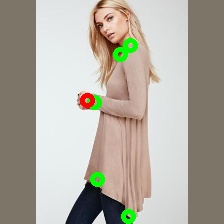}}
\subfloat[]{\includegraphics[width=0.20\textwidth]{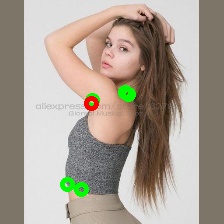}}
\subfloat[]{\includegraphics[width=0.20\textwidth]{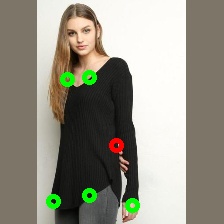}}\\
\subfloat[]{\includegraphics[width=0.20\textwidth]{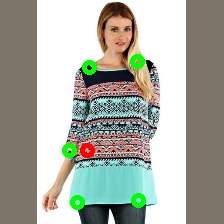}}
\subfloat[]{\includegraphics[width=0.20\textwidth]{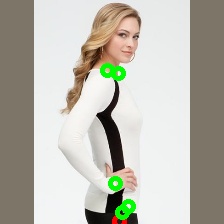}}
\subfloat[]{\includegraphics[width=0.20\textwidth]{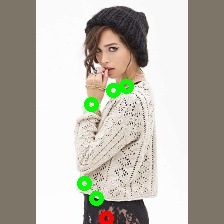}}
\subfloat[]{\includegraphics[width=0.20\textwidth]{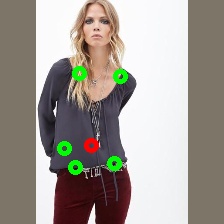}}
\subfloat[]{\includegraphics[width=0.20\textwidth]{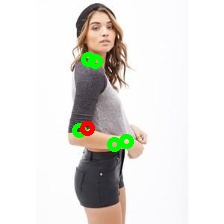}}  
      \caption{Example of images from the Fashion Landmark Dataset: landmarks detected as outliers by DeepGUM are shown 
in red, while inliers are shown in green. In all these images, the detected outliers correspond to occluded landmarks.}  
 
\label{fig:fashionVisible}
\end{figure*}

\begin{figure*}[p]
  \captionsetup[subfigure]{labelformat=empty}
  \centering
\subfloat[]{\includegraphics[width=0.20\textwidth]{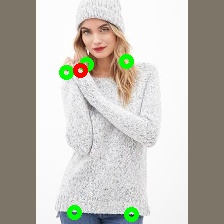}}
\subfloat[]{\includegraphics[width=0.20\textwidth]{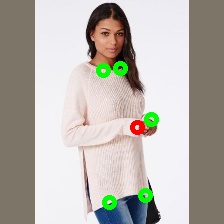}}
\subfloat[]{\includegraphics[width=0.20\textwidth]{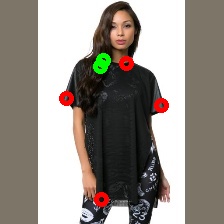}}
\subfloat[]{\includegraphics[width=0.20\textwidth]{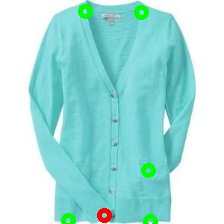}}
\subfloat[]{\includegraphics[width=0.20\textwidth]{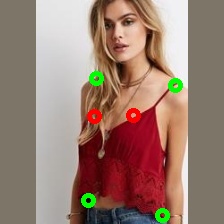}}\\
\subfloat[]{\includegraphics[width=0.20\textwidth]{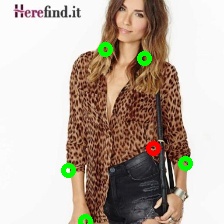}}
\subfloat[]{\includegraphics[width=0.20\textwidth]{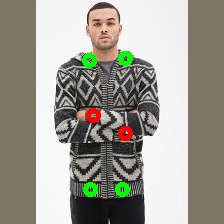}}
\subfloat[]{\includegraphics[width=0.20\textwidth]{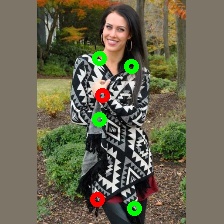}}
\subfloat[]{\includegraphics[width=0.20\textwidth]{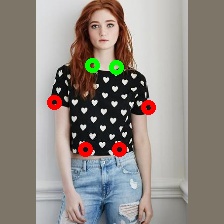}}
\subfloat[]{\includegraphics[width=0.20\textwidth]{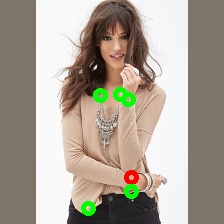}}\\
\subfloat[]{\includegraphics[width=0.20\textwidth]{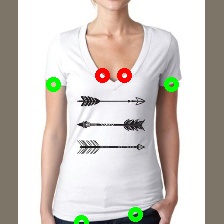}}
\subfloat[]{\includegraphics[width=0.20\textwidth]{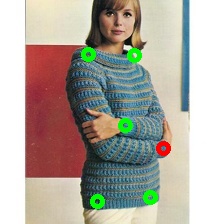}}
\subfloat[]{\includegraphics[width=0.20\textwidth]{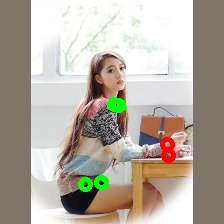}}
\subfloat[]{\includegraphics[width=0.20\textwidth]{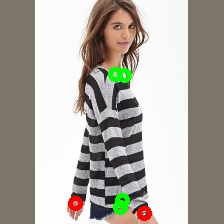}}
\subfloat[]{\includegraphics[width=0.20\textwidth]{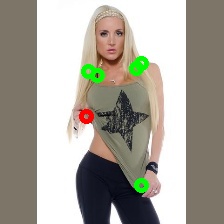}}\\
\subfloat[]{\includegraphics[width=0.20\textwidth]{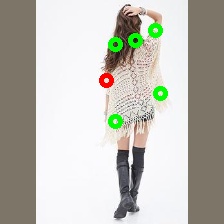}}
\subfloat[]{\includegraphics[width=0.20\textwidth]{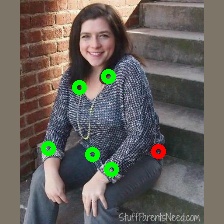}}
\subfloat[]{\includegraphics[width=0.20\textwidth]{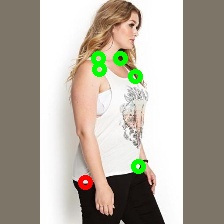}}
\subfloat[]{\includegraphics[width=0.20\textwidth]{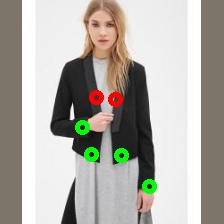}}
\subfloat[]{\includegraphics[width=0.20\textwidth]{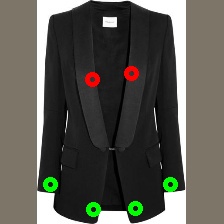}}\\
\subfloat[]{\includegraphics[width=0.20\textwidth]{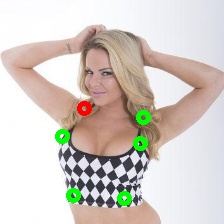}}
\subfloat[]{\includegraphics[width=0.20\textwidth]{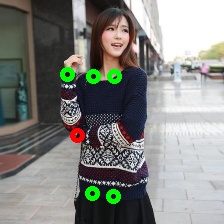}}
\subfloat[]{\includegraphics[width=0.20\textwidth]{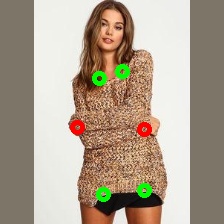}}
\subfloat[]{\includegraphics[width=0.20\textwidth]{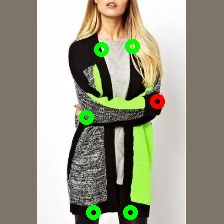}}
\subfloat[]{\includegraphics[width=0.20\textwidth]{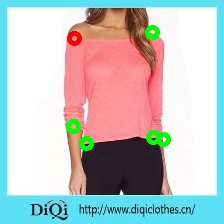}}
  \caption{Example of images from the Fashion Landmark Dataset: landmarks detected as outliers by DeepGUM are shown 
in red, while inliers are shown in green. The red landmarks correspond to inliers wrongly 
classified as outliers.}
\label{fig:fashionWrong}
\end{figure*}

In Section 4.1 of the manuscript, we presented experiments on the fashion landmark detection problem. In Figures~\ref{fig:fashionOK} 
to~\ref{fig:fashionWrong}, we show training examples containing at least one landmark that DeepGUM considers as 
outlier. These landmarks correspond to three different scenarios:
\begin{itemize}
 \item Figure~\ref{fig:fashionOK} shows images containing (i) either wrong annotations (e.g.\ last two images of the 
last row), (ii) ill-posed cases such as more than one clothe per image or (iii) challenging images (i.e.\ unusual 
clothing items like the third and last images of the fourth row).
 \item Figure~\ref{fig:fashionVisible} shows images in which one or more landmarks are visually occluded.
 \item Figure~\ref{fig:fashionWrong} shows images containing inlier landmarks wrongly classified as outliers by DeepGUM.
\end{itemize}

Table \ref{tab:results-fld2} displays results on two additional subsets of the fashion landmark detection dataset (full-body and lower-body landmarks). The results related to upper-body landmarks are already reported in the paper. Scores of DFA~[35] are not reported since authors do not provide results for these two subsets. These results confirm the superiority of the proposed model. Similarly to the first set (see the manuscript), DeepGUM performs best compared to the other robust methods and to $L_2$. This is again confirmed via statistical tests (except for the right hem in the full-body subset using $L_2$).

\begin{table*}[h]
\centering
\caption{Mean absolute error on the lower-body subsets of FLD, per landmark and in average. The landmarks are left (L) and right (R) hem (H) and trouser leg (T). DFA~[35] does not report on this subset. \label{tab:results-fld2-bis}\vspace{2mm}}
\begin{tabular}{llllll}
 \toprule
    \multirow{2}{*}{Method} & \multicolumn{5}{c}{Lower-body landmarks} \\
 \cmidrule{2-6} 
    & LH & RH & LT & RT & Avg. \\
 \midrule
    $L_2$ & 12.50 & 12.51 & 13.28 & 13.19 & 8.87 \\
    Huber~[36] & 22.98 & 14.42 & 43.76 & 58.47  &  31.45 \\
    Biweight~[30] & 15.75 & 15.77 & 18.00 & 17.96 & 11.99 \\
    DeepGUM & 12.19$^{***}$ & 12.23$^{***}$ & 12.80$^{***}$ & 12.81$^{***}$ & 8.53$^{***}$ \\
 \bottomrule
\end{tabular}
\end{table*}

\begin{table*}[h]
\centering
\caption{Mean absolute error on the full-body subset of FLD, per landmark and in average. The 
landmarks are left (L) and right (R) collar (C), sleeve (S), hem (H) and trouser leg 
(T). DFA~[35] does not report on this subset.\vspace{2mm}}
 \begin{tabular}{llllllllll}
 \toprule
 \multirow{2}{*}{Method} & \multicolumn{9}{c}{Full body landmarks} \\
 \cmidrule{2-10}
  & LC & RC & LS & RS & LH & RH & LT & RT & Avg. \\
 \midrule

  $L_2$ & 8.69 &  8.78 & 15.65 & 15.89 & 10.84 & 10.88$^{***}$ & 12.11 & 12.25 & 11.89 \\
  Huber~[36] & 14.26 &16.64  & 24.44 & 27.43 &19.14  & 21.28 & 26.99 &28.61  & 22.35 \\
  Biweight~[30] & 11.56 & 11.73 & 20.58 & 20.29 & 14.36 & 14.06 & 14.24 & 14.10 & 15.11 \\
  DeepGUM & 8.62$^{**}$ & 8.68$^{***}$ & 15.42$^{***}$ & 15.59$^{***}$ & 10.76$^{***}$ & 10.84$^{***}$ & 11.96$^{***}$ & 11.97$^{***}$ & 11.73$^{***}$ \\
  \bottomrule
 \end{tabular}
 \label{tab:results-fld2}
\end{table*}

%

\clearpage

\section{Age Estimation}

\begin{figure}[h!]
  \centering
  \includegraphics[width=0.90\columnwidth]{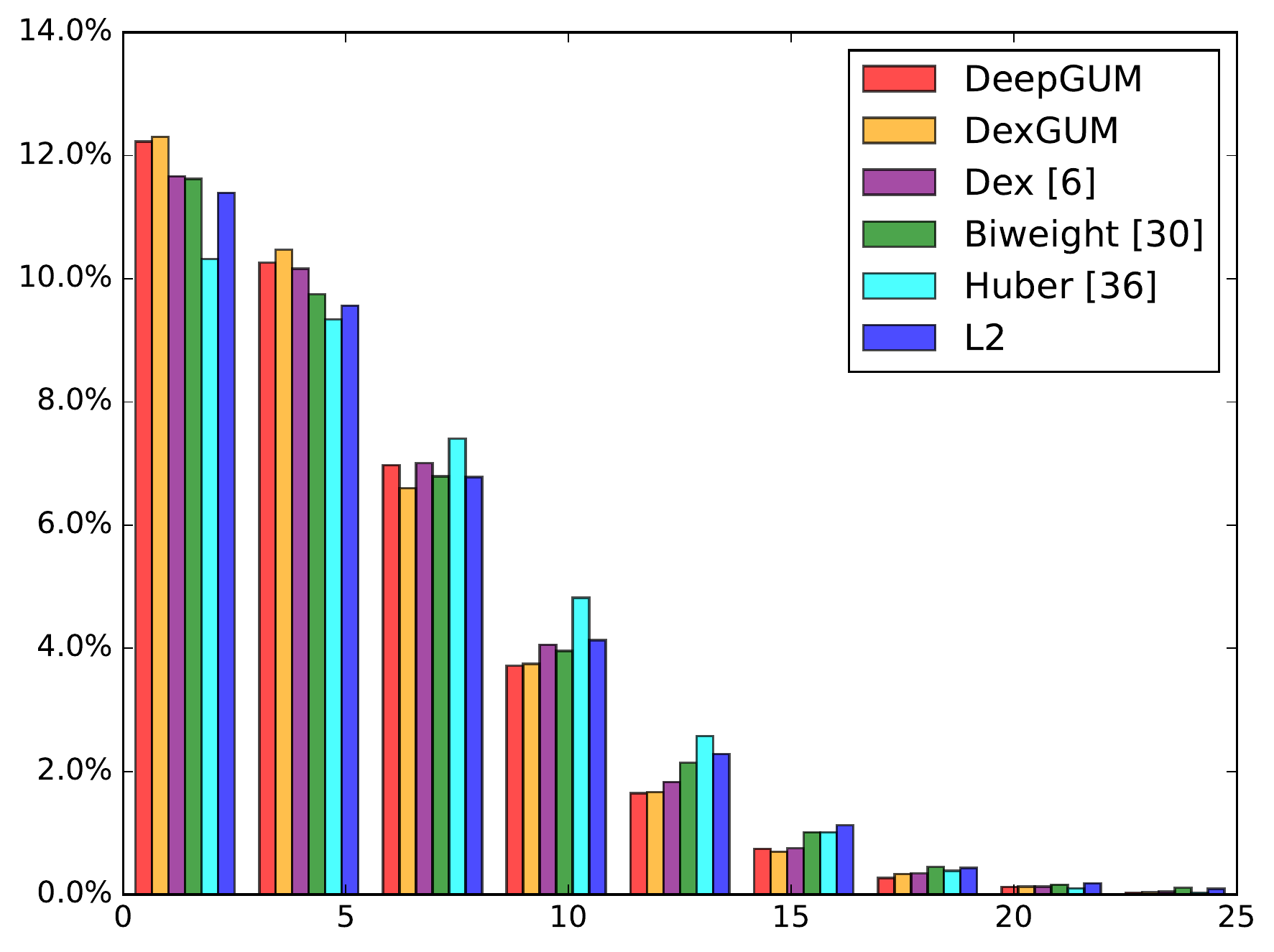}
    \caption{Histogram of the absolute error obtained with the different  methods tested on the CACD 
dataset.}
\label{fig:histAge}
\end{figure}

In Section 4.2, we presented experiments on the age estimation task. In Figure~\ref{fig:histAge}, we display the histogram of the absolute error obtained with different methods. We can see the importance of using DeepGUM to reduce the number of large errors. 

\begin{figure*}[p]
  \captionsetup[subfigure]{labelformat=empty}
  \centering
  \subfloat[55: 0.29]{\includegraphics[width=0.15\textwidth]{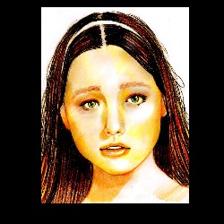}}
\subfloat[58: 0.29]{\includegraphics[width=0.15\textwidth]{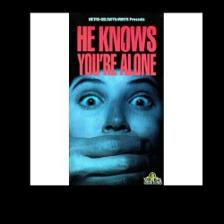}}
\subfloat[56: 0.28]{\includegraphics[width=0.15\textwidth]{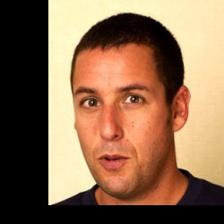}}
\subfloat[53: 0.30]{\includegraphics[width=0.15\textwidth]{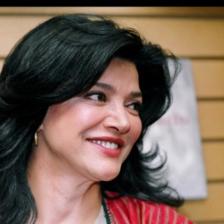}}
\subfloat[58: 0.25]{\includegraphics[width=0.15\textwidth]{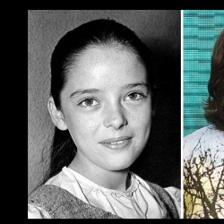}}
\subfloat[51: 0.31]{\includegraphics[width=0.15\textwidth]{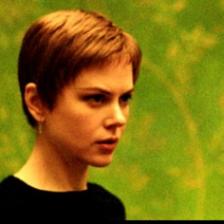}}\\
\subfloat[59: 0.31]{\includegraphics[width=0.15\textwidth]{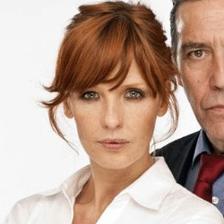}}
\subfloat[56: 0.29]{\includegraphics[width=0.15\textwidth]{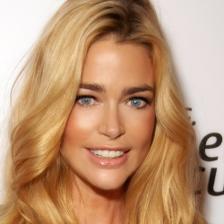}}
\subfloat[57: 0.30]{\includegraphics[width=0.15\textwidth]{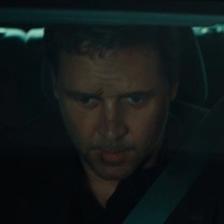}}
\subfloat[57: 0.25]{\includegraphics[width=0.15\textwidth]{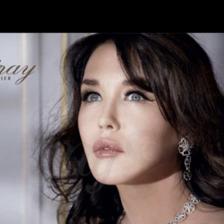}}
\subfloat[57: 0.31]{\includegraphics[width=0.15\textwidth]{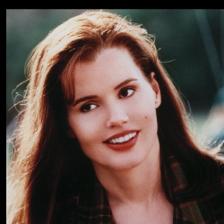}}
\subfloat[51: 0.29]{\includegraphics[width=0.15\textwidth]{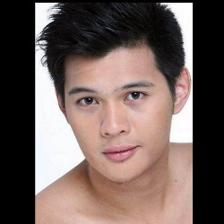}}\\
\subfloat[57: 0.25]{\includegraphics[width=0.15\textwidth]{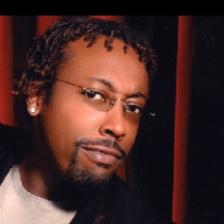}}
\subfloat[56: 0.28]{\includegraphics[width=0.15\textwidth]{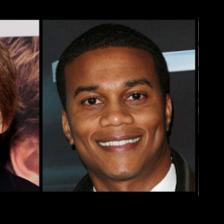}}
\subfloat[49: 0.20]{\includegraphics[width=0.15\textwidth]{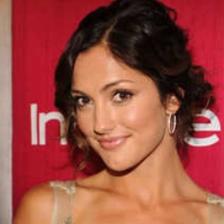}}
\subfloat[50: 0.32]{\includegraphics[width=0.15\textwidth]{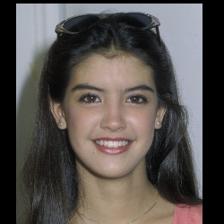}}
\subfloat[47: 0.31]{\includegraphics[width=0.15\textwidth]{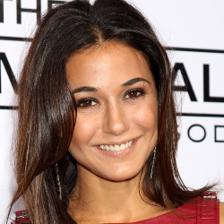}}
\subfloat[46: 0.31]{\includegraphics[width=0.15\textwidth]{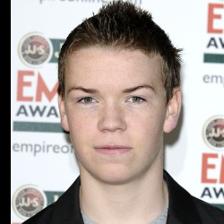}}\\
\subfloat[34: 0.31]{\includegraphics[width=0.15\textwidth]{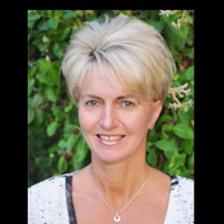}}
\subfloat[32: 0.25]{\includegraphics[width=0.15\textwidth]{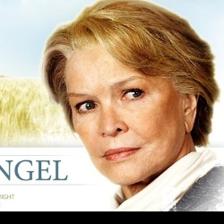}}
\subfloat[37: 0.26]{\includegraphics[width=0.15\textwidth]{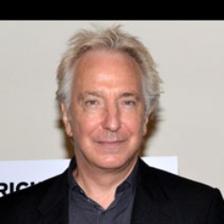}}
\subfloat[24: 0.30]{\includegraphics[width=0.15\textwidth]{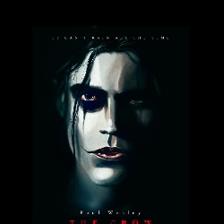}}
\subfloat[30: 0.20]{\includegraphics[width=0.15\textwidth]{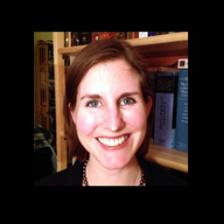}}
\subfloat[23: 0.28]{\includegraphics[width=0.15\textwidth]{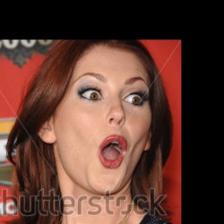}}\\
\subfloat[17: 0.31]{\includegraphics[width=0.15\textwidth]{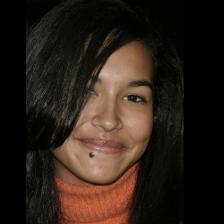}}
\subfloat[17: 0.31]{\includegraphics[width=0.15\textwidth]{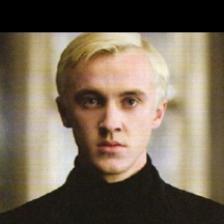}}
\subfloat[18: 0.25]{\includegraphics[width=0.15\textwidth]{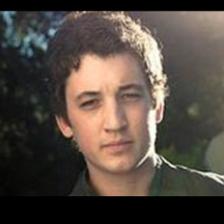}}
\subfloat[23: 0.32]{\includegraphics[width=0.15\textwidth]{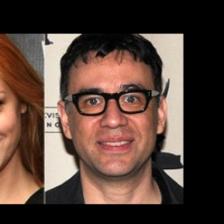}}
\subfloat[19: 0.25]{\includegraphics[width=0.15\textwidth]{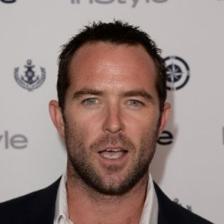}}
\subfloat[18: 0.27]{\includegraphics[width=0.15\textwidth]{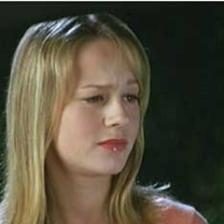}}\\
\subfloat[16: 0.25]{\includegraphics[width=0.15\textwidth]{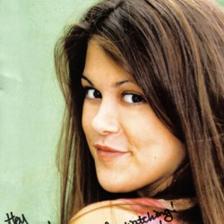}}
\subfloat[15: 0.27]{\includegraphics[width=0.15\textwidth]{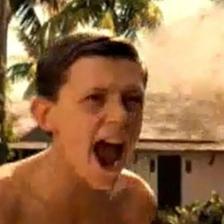}}
\subfloat[16: 0.31]{\includegraphics[width=0.15\textwidth]{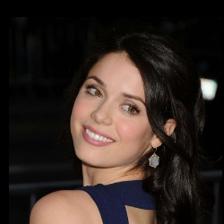}}
\subfloat[17: 0.31]{\includegraphics[width=0.15\textwidth]{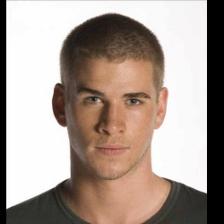}}
\subfloat[15: 0.31]{\includegraphics[width=0.15\textwidth]{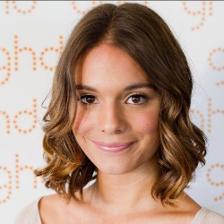}}
\subfloat[17: 0.26]{\includegraphics[width=0.15\textwidth]{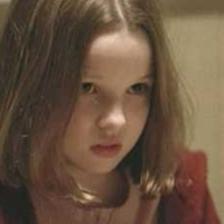}}\\
  \caption{Sample images of the CACD dataset estimated as outliers during training ($r_n<0.33$). The 
label below each image is the annotated age together with the $r_n$ at the end of the training of DeepGUM.}
\label{fig:033Age}
\end{figure*}

\begin{figure*}[p]
  \captionsetup[subfigure]{labelformat=empty}
  \centering
\subfloat[62: 0.58]{\includegraphics[width=0.15\textwidth]{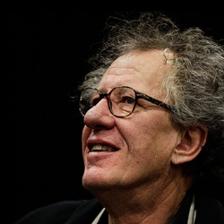}}
\subfloat[57: 0.55]{\includegraphics[width=0.15\textwidth]{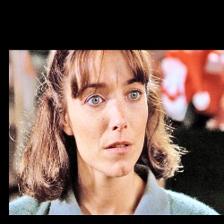}}
\subfloat[57: 0.36]{\includegraphics[width=0.15\textwidth]{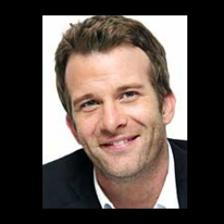}}
\subfloat[61: 0.51]{\includegraphics[width=0.15\textwidth]{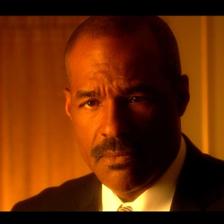}}
\subfloat[55: 0.62]{\includegraphics[width=0.15\textwidth]{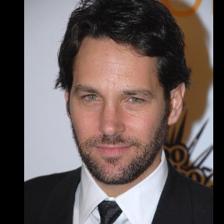}}
\subfloat[56: 0.46]{\includegraphics[width=0.15\textwidth]{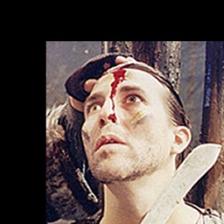}}\\
\subfloat[55: 0.41]{\includegraphics[width=0.15\textwidth]{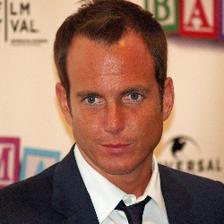}}
\subfloat[57: 0.51]{\includegraphics[width=0.15\textwidth]{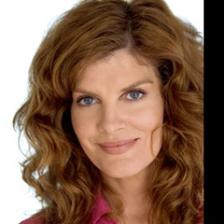}}
\subfloat[55: 0.61]{\includegraphics[width=0.15\textwidth]{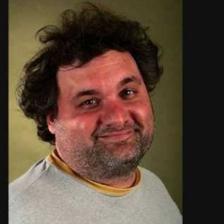}}
\subfloat[56: 0.62]{\includegraphics[width=0.15\textwidth]{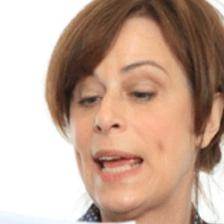}}
\subfloat[58: 0.36]{\includegraphics[width=0.15\textwidth]{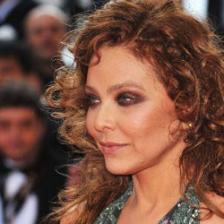}}
\subfloat[56: 0.34]{\includegraphics[width=0.15\textwidth]{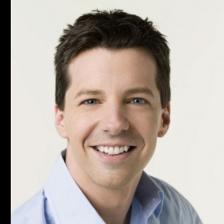}}\\
\subfloat[57: 0.62]{\includegraphics[width=0.15\textwidth]{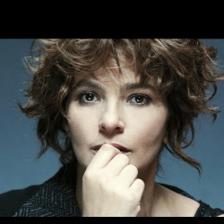}}
\subfloat[54: 0.58]{\includegraphics[width=0.15\textwidth]{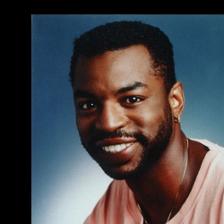}}
\subfloat[47: 0.58]{\includegraphics[width=0.15\textwidth]{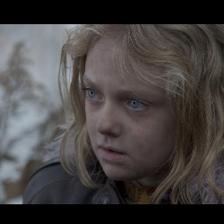}}
\subfloat[53: 0.51]{\includegraphics[width=0.15\textwidth]{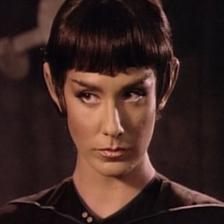}}
\subfloat[47: 0.65]{\includegraphics[width=0.15\textwidth]{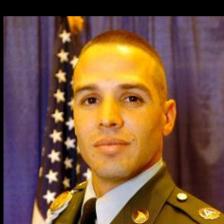}}
\subfloat[53: 0.64]{\includegraphics[width=0.15\textwidth]{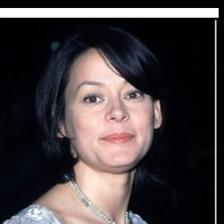}}\\
\subfloat[50: 0.34]{\includegraphics[width=0.15\textwidth]{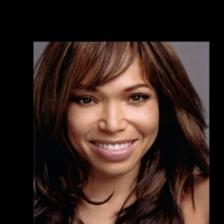}}
\subfloat[51: 0.42]{\includegraphics[width=0.15\textwidth]{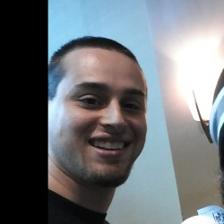}}
\subfloat[47: 0.65]{\includegraphics[width=0.15\textwidth]{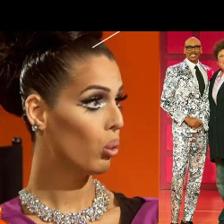}}
\subfloat[39: 0.59]{\includegraphics[width=0.15\textwidth]{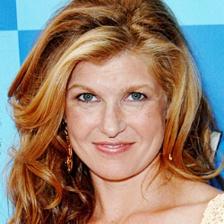}}
\subfloat[40: 0.60]{\includegraphics[width=0.15\textwidth]{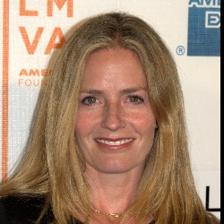}}
\subfloat[33: 0.49]{\includegraphics[width=0.15\textwidth]{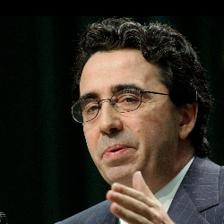}}\\
\subfloat[30: 0.59]{\includegraphics[width=0.15\textwidth]{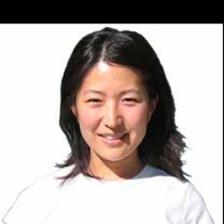}}
\subfloat[27: 0.61]{\includegraphics[width=0.15\textwidth]{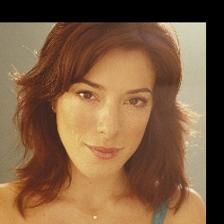}}
\subfloat[27: 0.52]{\includegraphics[width=0.15\textwidth]{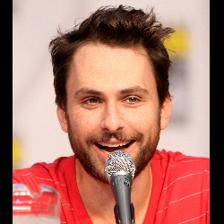}}
\subfloat[27: 0.40]{\includegraphics[width=0.15\textwidth]{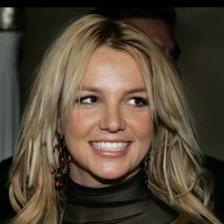}}
\subfloat[24: 0.53]{\includegraphics[width=0.15\textwidth]{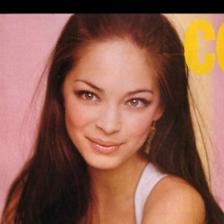}}
\subfloat[24: 0.64]{\includegraphics[width=0.15\textwidth]{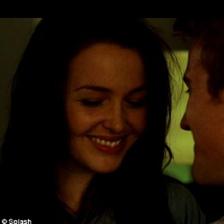}}\\
\subfloat[21: 0.49]{\includegraphics[width=0.15\textwidth]{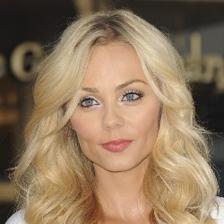}}
\subfloat[22: 0.60]{\includegraphics[width=0.15\textwidth]{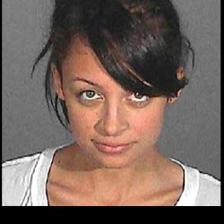}}
\subfloat[20: 0.52]{\includegraphics[width=0.15\textwidth]{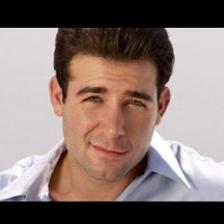}}
\subfloat[22: 0.65]{\includegraphics[width=0.15\textwidth]{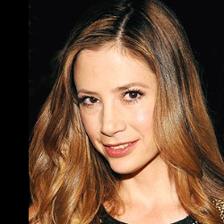}}
\subfloat[20: 0.46]{\includegraphics[width=0.15\textwidth]{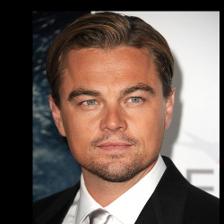}}
\subfloat[21: 0.61]{\includegraphics[width=0.15\textwidth]{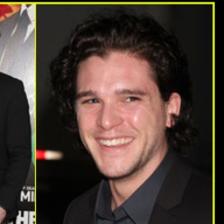}}

  \caption{Sample images of the CACD dataset with high outlier uncertainty ($0.33<r_n<0.66$). The 
label below each image is the annotated age together with the $r_n$ at the end of the training of DeepGUM.}
\label{fig:033_066Age}
\end{figure*}

\begin{figure*}[p]
  \captionsetup[subfigure]{labelformat=empty}
  \centering
\subfloat[53: 0.95]{\includegraphics[width=0.15\textwidth]{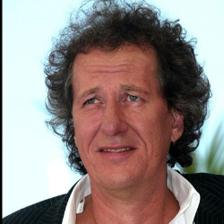}}
\subfloat[57: 0.80]{\includegraphics[width=0.15\textwidth]{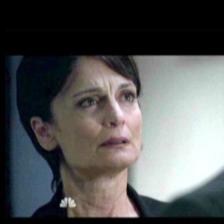}}
\subfloat[57: 0.94]{\includegraphics[width=0.15\textwidth]{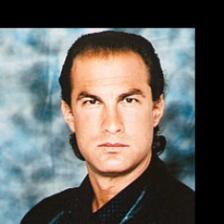}}
\subfloat[58: 0.95]{\includegraphics[width=0.15\textwidth]{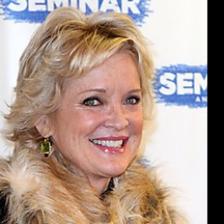}}
\subfloat[53: 0.92]{\includegraphics[width=0.15\textwidth]{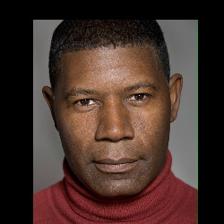}}
\subfloat[54: 0.94]{\includegraphics[width=0.15\textwidth]{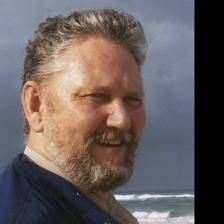}}\\
\subfloat[55: 0.87]{\includegraphics[width=0.15\textwidth]{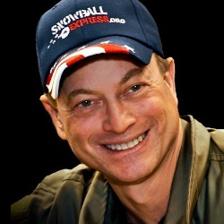}}
\subfloat[50: 0.95]{\includegraphics[width=0.15\textwidth]{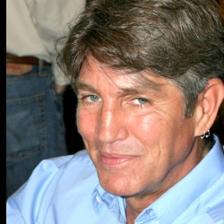}}
\subfloat[54: 0.95]{\includegraphics[width=0.15\textwidth]{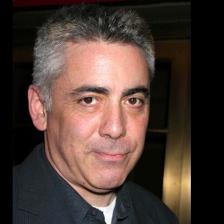}}
\subfloat[48: 0.95]{\includegraphics[width=0.15\textwidth]{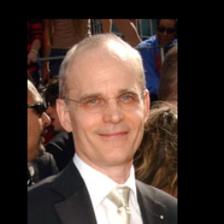}}
\subfloat[47: 0.94]{\includegraphics[width=0.15\textwidth]{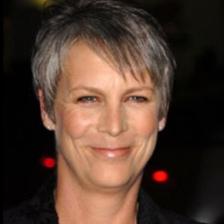}}
\subfloat[42: 0.93]{\includegraphics[width=0.15\textwidth]{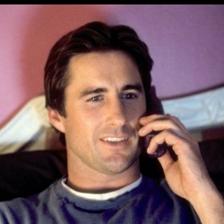}}\\
\subfloat[50: 0.94]{\includegraphics[width=0.15\textwidth]{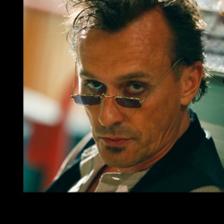}}
\subfloat[48: 0.94]{\includegraphics[width=0.15\textwidth]{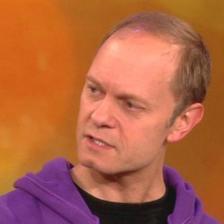}}
\subfloat[46: 0.95]{\includegraphics[width=0.15\textwidth]{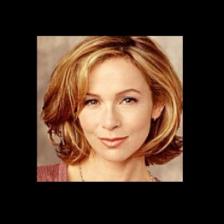}}
\subfloat[45: 0.95]{\includegraphics[width=0.15\textwidth]{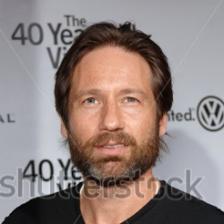}}
\subfloat[50: 0.95]{\includegraphics[width=0.15\textwidth]{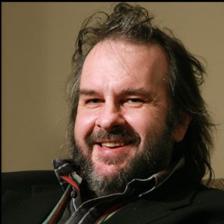}}
\subfloat[44: 0.95]{\includegraphics[width=0.15\textwidth]{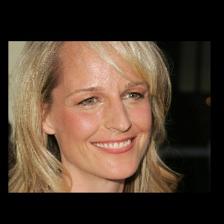}}\\
\subfloat[45: 0.94]{\includegraphics[width=0.15\textwidth]{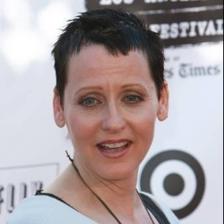}}
\subfloat[46: 0.95]{\includegraphics[width=0.15\textwidth]{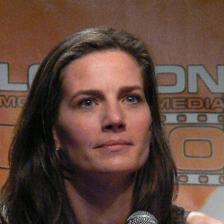}}
\subfloat[47: 0.95]{\includegraphics[width=0.15\textwidth]{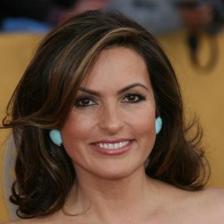}}
\subfloat[46: 0.91]{\includegraphics[width=0.15\textwidth]{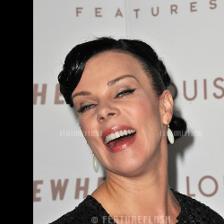}}
\subfloat[43: 0.93]{\includegraphics[width=0.15\textwidth]{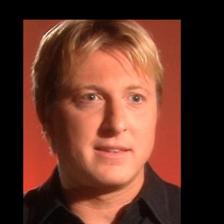}}
\subfloat[43: 0.95]{\includegraphics[width=0.15\textwidth]{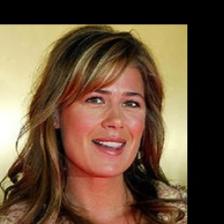}}\\
\subfloat[41: 0.95]{\includegraphics[width=0.15\textwidth]{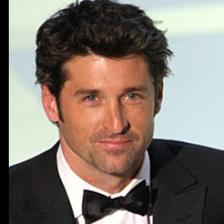}}
\subfloat[38: 0.95]{\includegraphics[width=0.15\textwidth]{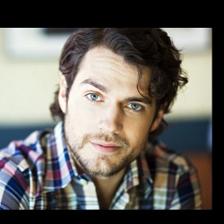}}
\subfloat[37: 0.94]{\includegraphics[width=0.15\textwidth]{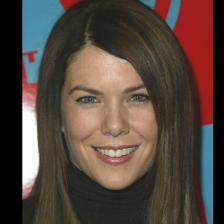}}
\subfloat[44: 0.95]{\includegraphics[width=0.15\textwidth]{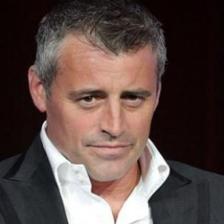}}
\subfloat[43: 0.95]{\includegraphics[width=0.15\textwidth]{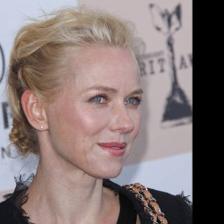}}
\subfloat[38: 0.95]{\includegraphics[width=0.15\textwidth]{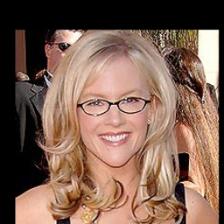}}\\
\subfloat[43: 0.89]{\includegraphics[width=0.15\textwidth]{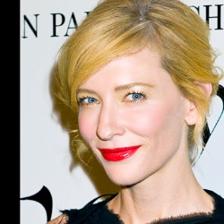}}
\subfloat[42: 0.93]{\includegraphics[width=0.15\textwidth]{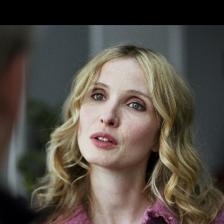}}
\subfloat[39: 0.95]{\includegraphics[width=0.15\textwidth]{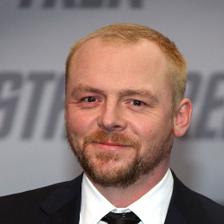}}
\subfloat[41: 0.95]{\includegraphics[width=0.15\textwidth]{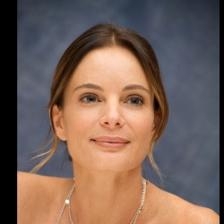}}
\subfloat[42: 0.93]{\includegraphics[width=0.15\textwidth]{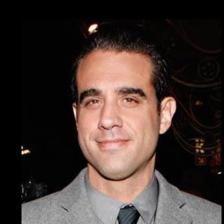}}
\subfloat[38: 0.91]{\includegraphics[width=0.15\textwidth]{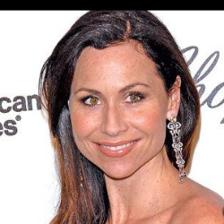}}\\

  \caption{Sample images of the CACD dataset estimated as inliers during training ($r_n>0.66$). The 
label below each image is the annotated age together with the $r_n$ at the end of the DeepGUM training.}
\label{fig:066Age}
\end{figure*}

Figures from~\ref{fig:033Age} to~\ref{fig:066Age} display three different groups of images depending on the probability of being inlier that DeepGUM assigns to each age annotation:
\begin{itemize}
 \item Figure~\ref{fig:033Age} shows randomly selected images with a high probability of being outliers according to 
DeepGUM ($r_n<0.33$). Even if some of the results could be debatable, we argue that most of the annotations (displayed 
below the image) are incorrect. DeepGUM correctly performs the task for which it was designed.
 \item Figure~\ref{fig:033_066Age} displays randomly selected images for which the network has trouble deciding between 
inlier and outlier ($0.33<r_n<0.66$). Even more, for most of these images it is quite hard to decide whether the 
annotation is correct or not.
 \item Figure~\ref{fig:066Age} shows randomly selected images that are considered by DeepGUM
as inliers ($0.66<r_n$). Indeed, the annotation below each image looks correct in most of the cases.
\end{itemize}

\clearpage
\section{Head pose estimation}

\begin{figure*}[h!]
  \centering
  \includegraphics[width=1\textwidth]{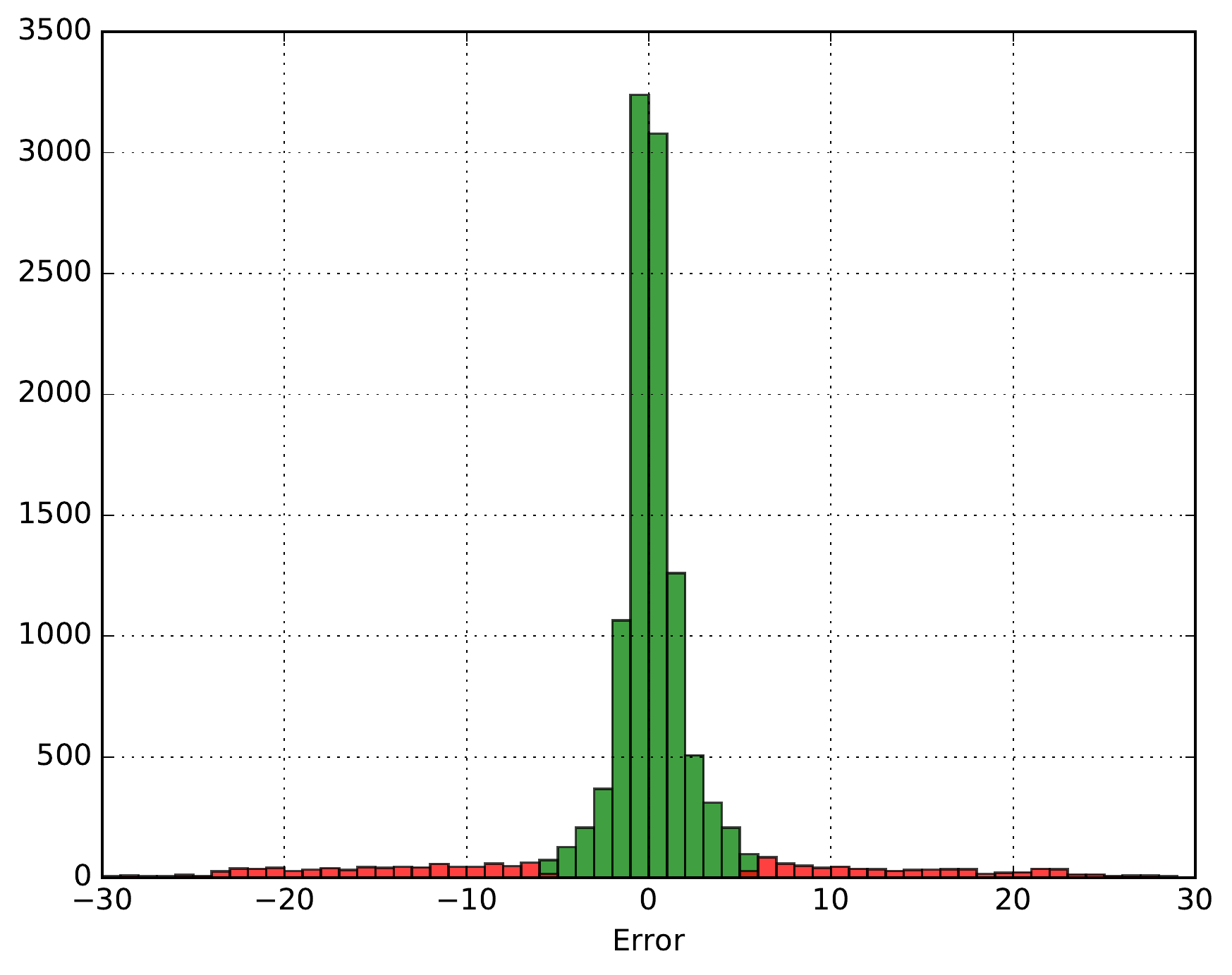}
  \caption{Error histogram on the McGill Dataset. Points that are considered as outliers are displayed in red ($r_n<0.5$) and inliers are displayed in green ($r_n\geq0.5$)}
\label{fig:erroMCGill}
\end{figure*}

\begin{figure*}[p]
  \captionsetup[subfigure]{labelformat=empty}
  \centering
\subfloat[-68]{\includegraphics[width=0.15\textwidth]{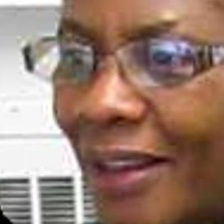}}
\subfloat[23 ]{\includegraphics[width=0.15\textwidth]{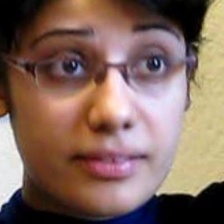}}
\subfloat[-90]{\includegraphics[width=0.15\textwidth]{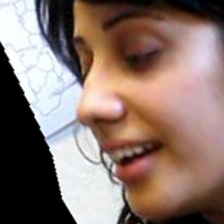}}
\subfloat[23 ]{\includegraphics[width=0.15\textwidth]{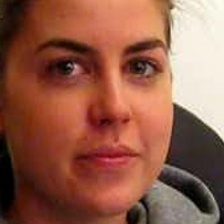}}
\subfloat[0  ]{\includegraphics[width=0.15\textwidth]{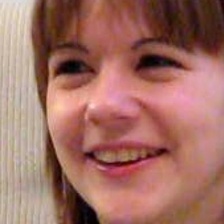}}
\subfloat[-45]{\includegraphics[width=0.15\textwidth]{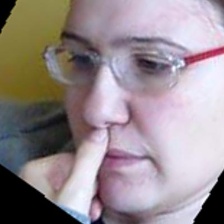}}\\
\subfloat[0  ]{\includegraphics[width=0.15\textwidth]{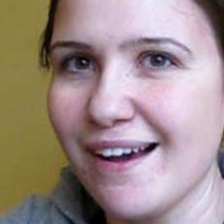}}
\subfloat[0  ]{\includegraphics[width=0.15\textwidth]{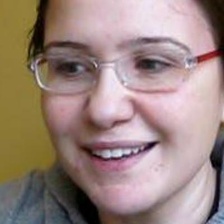}}
\subfloat[0  ]{\includegraphics[width=0.15\textwidth]{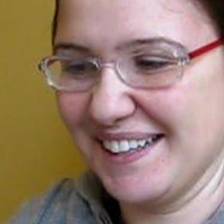}}
\subfloat[68 ]{\includegraphics[width=0.15\textwidth]{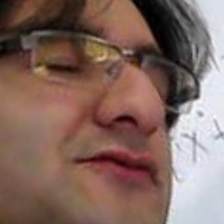}}
\subfloat[0  ]{\includegraphics[width=0.15\textwidth]{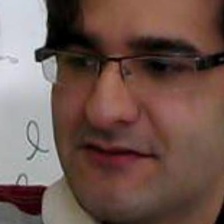}}
\subfloat[23 ]{\includegraphics[width=0.15\textwidth]{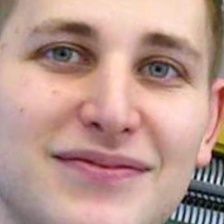}}\\
\subfloat[23 ]{\includegraphics[width=0.15\textwidth]{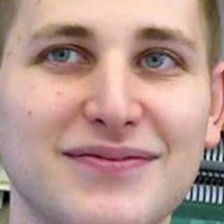}}
\subfloat[0  ]{\includegraphics[width=0.15\textwidth]{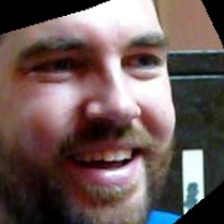}}
\subfloat[0  ]{\includegraphics[width=0.15\textwidth]{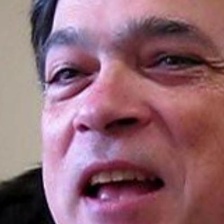}}
\subfloat[23 ]{\includegraphics[width=0.15\textwidth]{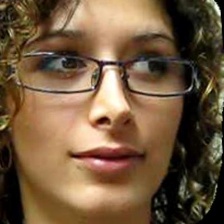}}
\subfloat[23 ]{\includegraphics[width=0.15\textwidth]{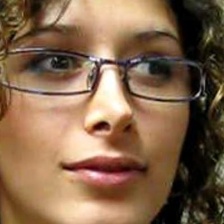}}
\subfloat[-90]{\includegraphics[width=0.15\textwidth]{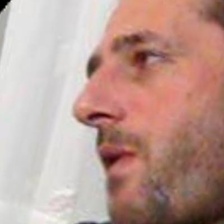}}\\
\subfloat[-90]{\includegraphics[width=0.15\textwidth]{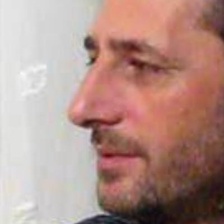}}
\subfloat[-90]{\includegraphics[width=0.15\textwidth]{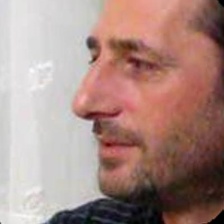}}
\subfloat[-45]{\includegraphics[width=0.15\textwidth]{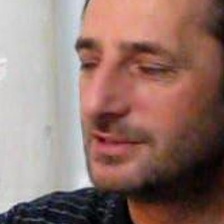}}
\subfloat[0  ]{\includegraphics[width=0.15\textwidth]{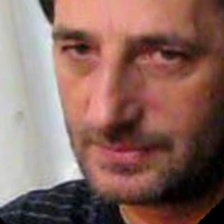}}
\subfloat[-45]{\includegraphics[width=0.15\textwidth]{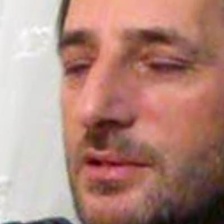}}
\subfloat[-45]{\includegraphics[width=0.15\textwidth]{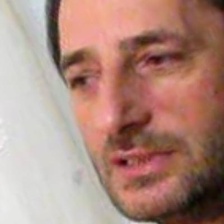}}\\
\subfloat[-45]{\includegraphics[width=0.15\textwidth]{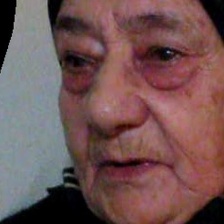}}
\subfloat[-45]{\includegraphics[width=0.15\textwidth]{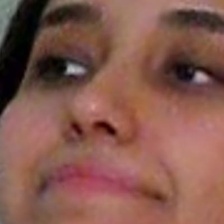}}
\subfloat[23 ]{\includegraphics[width=0.15\textwidth]{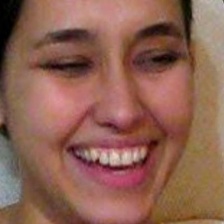}}
\subfloat[-45]{\includegraphics[width=0.15\textwidth]{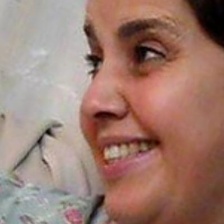}}
\subfloat[-90]{\includegraphics[width=0.15\textwidth]{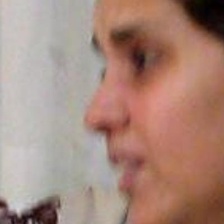}}
\subfloat[-90]{\includegraphics[width=0.15\textwidth]{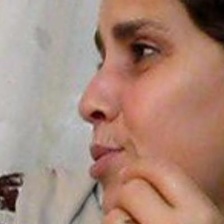}}\\
\subfloat[-45]{\includegraphics[width=0.15\textwidth]{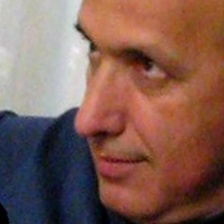}}
\subfloat[68 ]{\includegraphics[width=0.15\textwidth]{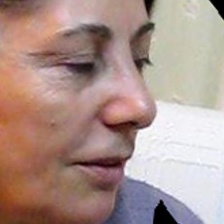}}
\subfloat[90 ]{\includegraphics[width=0.15\textwidth]{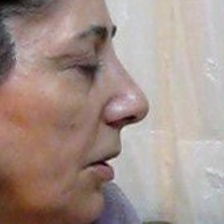}}
\subfloat[0  ]{\includegraphics[width=0.15\textwidth]{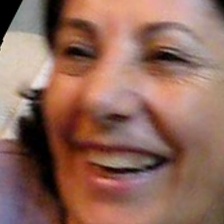}}
\subfloat[23 ]{\includegraphics[width=0.15\textwidth]{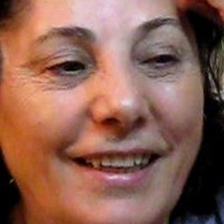}}
\subfloat[23 ]{\includegraphics[width=0.15\textwidth]{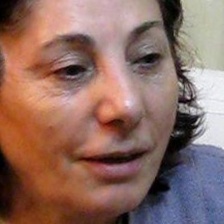}}

  \caption{Sample images from the McGill dataset considered as outliers during training ($r_n<10^{-4}$). The label below each image is the angle included in the annotation.}
\label{fig:HPE}
\end{figure*}

In Section 4.3 of the manuscript, we presented experiments on the head pose estimation task. We illustrated the 
benefit of our propoal that robustly detect outliers at training time, Figure~\ref{fig:HPE} shows images from the McGill dataset the DeepGUM considered 
as outlier. In these examples, many clear outliers appear. For some images, it is difficult to say if the annotation is good even for a human annotator.
In Figure~\ref{fig:erroMCGill}, we display the error obtained on one fold of the training set. It visually justifies the choice of a Gaussian-Uniform model for the error distribution.

\clearpage
\section{Facial Landmark Detection}
Section 4.4 of the manuscript reports experiments on the facial landmark detection (FLD) task. We showed the 
benefit of using DeepGUM, a robust regression approach able to detect outliers at training time, under the 
presence of different kinds of corrupting outliers. Figure~\ref{fig:Facial} shows images from FLD corrupted with the 
\emph{l-UGO} strategy and $30\%$ of outliers (thus these images correspond to the point of the red curve in Figure 5.a in the main manuscript with x-axis value equal to $30\%$). Superposed to the images we can see circles and crosses for DeepGUM and Biweight respectively, located at the (corrupted) annotations. Since in this dataset the closest competitor is Biweight and plotting the results of more than two methods per image would be unintelligible, we do not report results for Huber. The color indicates whether each method detects the annotated landmark as an outlier (red) or inliner (green). In the case of Biweight, since the method is acts coordinate-wise, there are vertical and horizontal lines, denoting whether the vertical or horizontal coordinates respectively are detected as outliers. First of all we remark that almost all of the uncorrupted landmarks are detected as inliers by both methods. This corresponds to the 100\% outlier precision (i.e. no inliers are classified as outliers) in the curves of Figure~5.a. Regarding the detection of outliers, we can see that Biweight classifies many outliers as inliers (lower outlier recall with respect to DeepGUM, as in Figure~5.a). We can also observe that, because Biweight works coordinate-wise, some of the landmarks are detected as outliers horizontally and not vertically and vice-versa. For instance in the first row fifth column, we can see that the nose landmark is wrongly annotated as close to the eyebrow. Horizontally the error is not big, and therefore Biweight classifies this as a horizontal inlier and vertical outlier. However, this is wrong, because ideally we would not want to use the eyebrow as a nose samples. Other examples confirm this behavior and explain, not only why the recall of Biweight is lower than DeepGUM, but also one of the reasons of the difference in performance.

\begin{figure}[H]
\centering
\includegraphics[width=\linewidth]{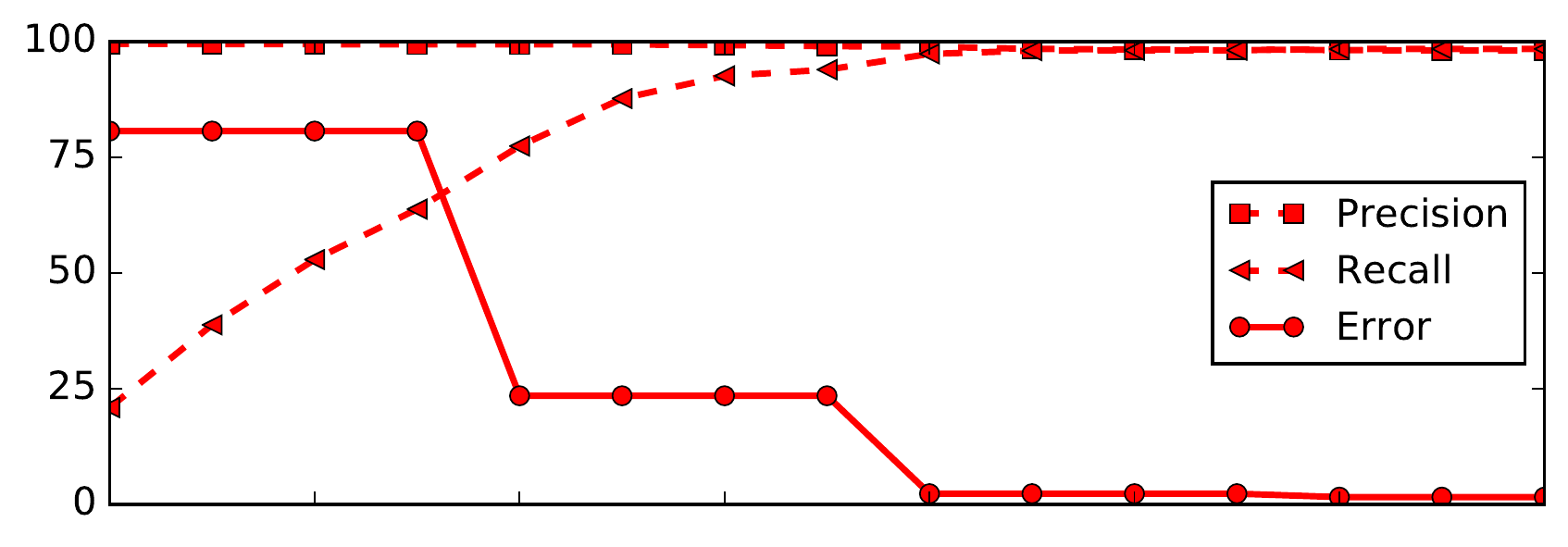}
\caption{Precision and recall (on the training set) as MAE (on the test set) over the M-step iterations with \textit{l-UGO} noise at $30\%$.\label{fig:evolution}}
\end{figure}

\begin{figure*}[p]
  \captionsetup[subfigure]{labelformat=empty}
  \centering
\subfloat[]{\includegraphics[width=0.15\textwidth]{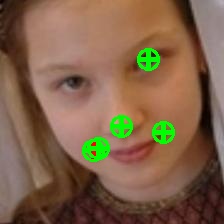}}
\subfloat[]{\includegraphics[width=0.15\textwidth]{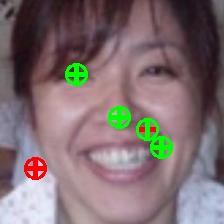}}
\subfloat[]{\includegraphics[width=0.15\textwidth]{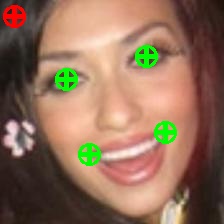}}
\subfloat[]{\includegraphics[width=0.15\textwidth]{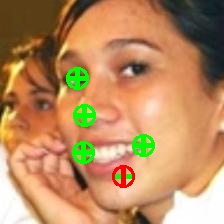}}
\subfloat[]{\includegraphics[width=0.15\textwidth]{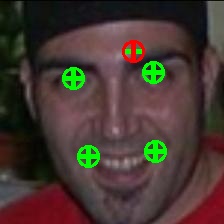}}
\subfloat[]{\includegraphics[width=0.15\textwidth]{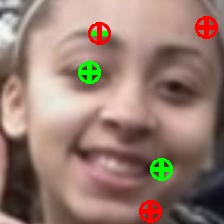}}\\
\vspace{-0.5cm}
\subfloat[]{\includegraphics[width=0.15\textwidth]{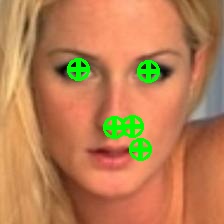}}
\subfloat[]{\includegraphics[width=0.15\textwidth]{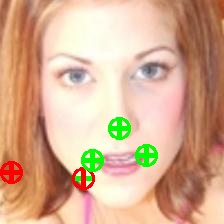}}
\subfloat[]{\includegraphics[width=0.15\textwidth]{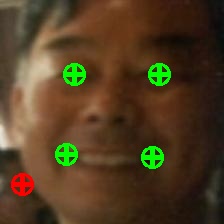}}
\subfloat[]{\includegraphics[width=0.15\textwidth]{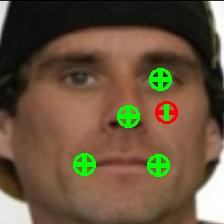}}
\subfloat[]{\includegraphics[width=0.15\textwidth]{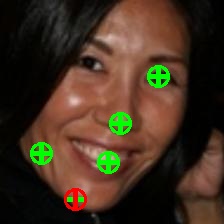}}
\subfloat[]{\includegraphics[width=0.15\textwidth]{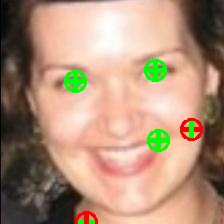}}\\
\vspace{-0.5cm}
\subfloat[]{\includegraphics[width=0.15\textwidth]{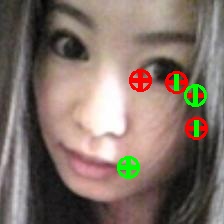}}
\subfloat[]{\includegraphics[width=0.15\textwidth]{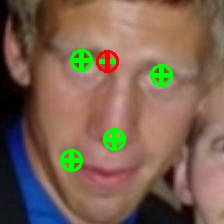}}
\subfloat[]{\includegraphics[width=0.15\textwidth]{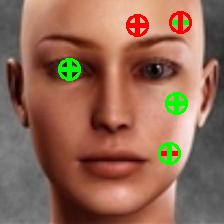}}
\subfloat[]{\includegraphics[width=0.15\textwidth]{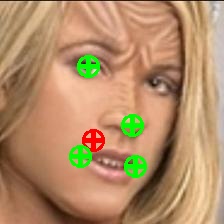}}
\subfloat[]{\includegraphics[width=0.15\textwidth]{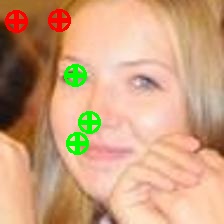}}
\subfloat[]{\includegraphics[width=0.15\textwidth]{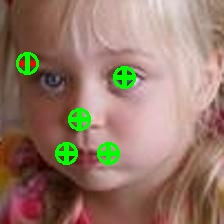}}\\
\vspace{-0.5cm}
\subfloat[]{\includegraphics[width=0.15\textwidth]{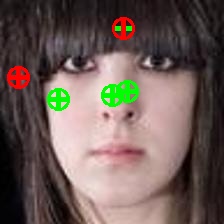}}
\subfloat[]{\includegraphics[width=0.15\textwidth]{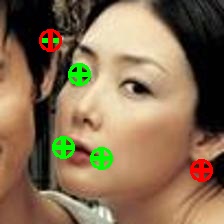}}
\subfloat[]{\includegraphics[width=0.15\textwidth]{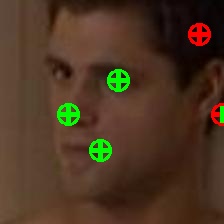}}
\subfloat[]{\includegraphics[width=0.15\textwidth]{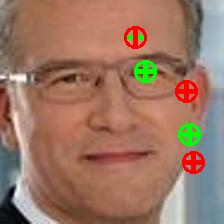}}
\subfloat[]{\includegraphics[width=0.15\textwidth]{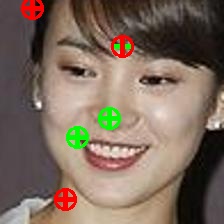}}
\subfloat[]{\includegraphics[width=0.15\textwidth]{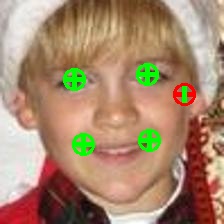}}\\
\vspace{-0.5cm}
\subfloat[]{\includegraphics[width=0.15\textwidth]{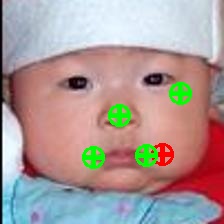}}
\subfloat[]{\includegraphics[width=0.15\textwidth]{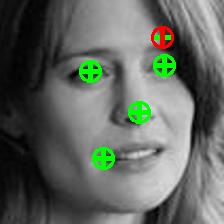}}
\subfloat[]{\includegraphics[width=0.15\textwidth]{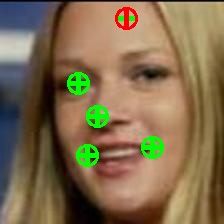}}
\subfloat[]{\includegraphics[width=0.15\textwidth]{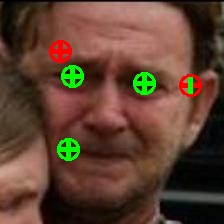}}
\subfloat[]{\includegraphics[width=0.15\textwidth]{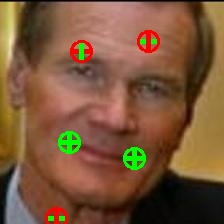}}
\subfloat[]{\includegraphics[width=0.15\textwidth]{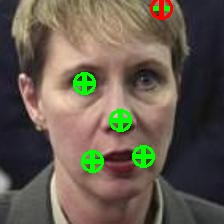}}\\
\vspace{-0.5cm}
\subfloat[]{\includegraphics[width=0.15\textwidth]{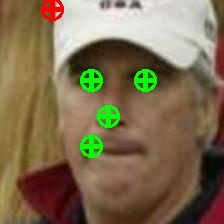}}
\subfloat[]{\includegraphics[width=0.15\textwidth]{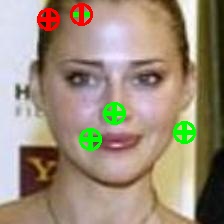}}
\subfloat[]{\includegraphics[width=0.15\textwidth]{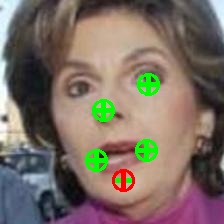}}
\subfloat[]{\includegraphics[width=0.15\textwidth]{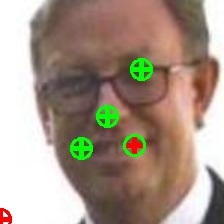}}
\subfloat[]{\includegraphics[width=0.15\textwidth]{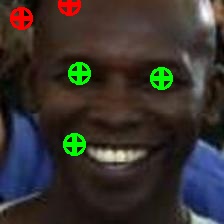}}
\subfloat[]{\includegraphics[width=0.15\textwidth]{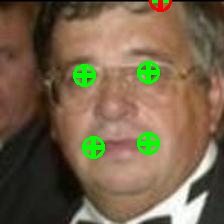}}\\
\vspace{-0.5cm}
\subfloat[]{\includegraphics[width=0.15\textwidth]{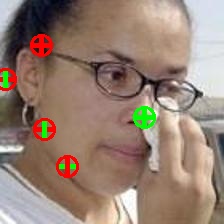}}
\subfloat[]{\includegraphics[width=0.15\textwidth]{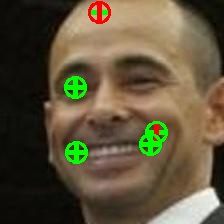}}
\subfloat[]{\includegraphics[width=0.15\textwidth]{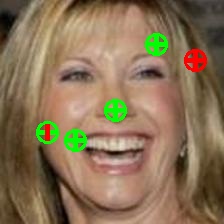}}
\subfloat[]{\includegraphics[width=0.15\textwidth]{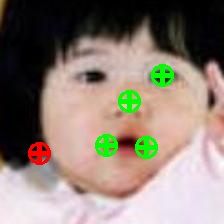}}
\subfloat[]{\includegraphics[width=0.15\textwidth]{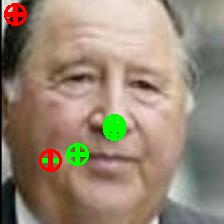}}
\subfloat[]{\includegraphics[width=0.15\textwidth]{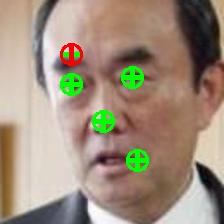}}

\caption{Sample images of the facial landmark detection problem: landmarks that are considered as outliers during training are displayed in red while the ones classified as inliers are displayed in green. The results of DeepGUM are displayed with circles, while those of Biweight are displayed in vertical and horizontal lines (independent detections per coordinate).}
\label{fig:Facial}
\end{figure*}

In order to better understand the training procedure of DeepGUM, we plot three curves in Figure~\ref{fig:evolution}. These curves represent the same quantities as in Figure~5, but the x-axis takes a different meaning in this plot. Indeed, two of these curves correspond to the precision (squares) and recall (triangles), both dashed, of the training set. The third curve (circles-solid) corresponds to the MAE of the test set (which is clean). However, the abscissa correspond to the M-step iterations, i.e.\ update of the parameters of the graphical model, $\thetavect$, using equations~(3), (4) and~(5). When the test error is flat, the EM is looping and therefore the network weights $\wvect$ are not updated (so the test set is constant). Once the EM converged, SGD takes over until convergence and a new execution over the M-step (with the new network weights and hopefully lower test error). We can see that the recall is increasing progressively, even when the network weights are constant, meaning that the EM is actually discovering in a progressively more efficient manner the outliers in the training set.\vspace{-4mm}

%

\end{document}